\documentclass[letterpaper]{article}
\usepackage{aaai24}
\usepackage{times}
\usepackage{helvet}
\usepackage{courier}
\usepackage[hyphens]{url}
\usepackage{graphicx}
\def\UrlFont{\rm}
\usepackage{natbib}
\usepackage{caption}
\usepackage{subcaption}
\usepackage{blindtext}
\usepackage{multirow}
\usepackage{amsmath}
\usepackage{amssymb}
\usepackage[dvipsnames]{xcolor}

\usepackage{hyperref}
\usepackage{makecell}
\usepackage{algorithm}
\usepackage{algorithmic}
\usepackage{newfloat}
\usepackage{listings}
\DeclareCaptionStyle{ruled}{labelfont=normalfont,labelsep=colon,strut=off}
\floatstyle{ruled}
\newfloat{listing}{tb}{lst}{}
\floatname{listing}{Listing}

\author{
    Namhyuk Ahn\textsuperscript{\rm 1}\quad
    Junsoo Lee\textsuperscript{\rm 1}\quad
    Chunggi Lee\textsuperscript{\rm 1,2}\\
    Kunhee Kim\textsuperscript{\rm 3}\quad
    Daesik Kim\textsuperscript{\rm 1}\quad
    Seung-Hun Nam\textsuperscript{\rm 1}\quad
    Kibeom Hong\textsuperscript{\rm 4}
}
\affiliations{
    \textsuperscript{\rm 1} NAVER WEBTOON AI\quad
    \textsuperscript{\rm 2} Harvard University\quad
    \textsuperscript{\rm 3} KAIST\quad
    \textsuperscript{\rm 4} SwatchOn
}

\DeclareMathOperator*{\argmin}{arg\,min}
\newcommand{\ours}{DreamStyler}

\newcommand{\eref}[1]{Eq. \eqref{#1}}
\newcommand{\tref}[1]{Table \ref{#1}}
\newcommand{\fref}[1]{Figure \ref{#1}}

\title{\ours: Paint by Style Inversion with Text-to-Image Diffusion Models}

\begin{document}

\maketitle

\newcommand{\figTeaser}{
\begin{figure}[t]
\centering
\includegraphics[width=\linewidth]{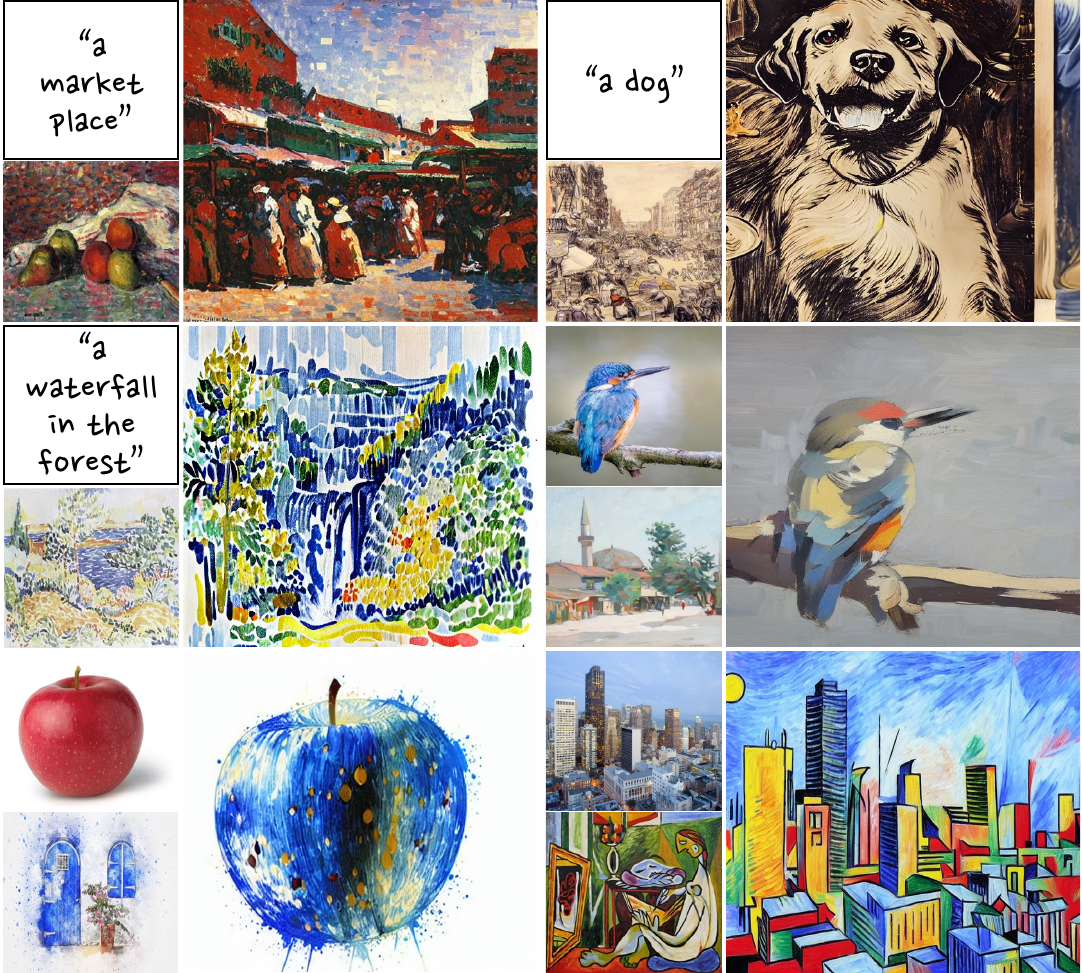}
\caption{\textbf{\ours} synthesizes outputs based on a given context along with a style reference. Note that each model is trained on a single style image shown in this figure.}
\label{fig:teaser}
\end{figure}
}

\newcommand{\figModelOverview}{
\begin{figure}[t]
\centering
\includegraphics[width=\linewidth]{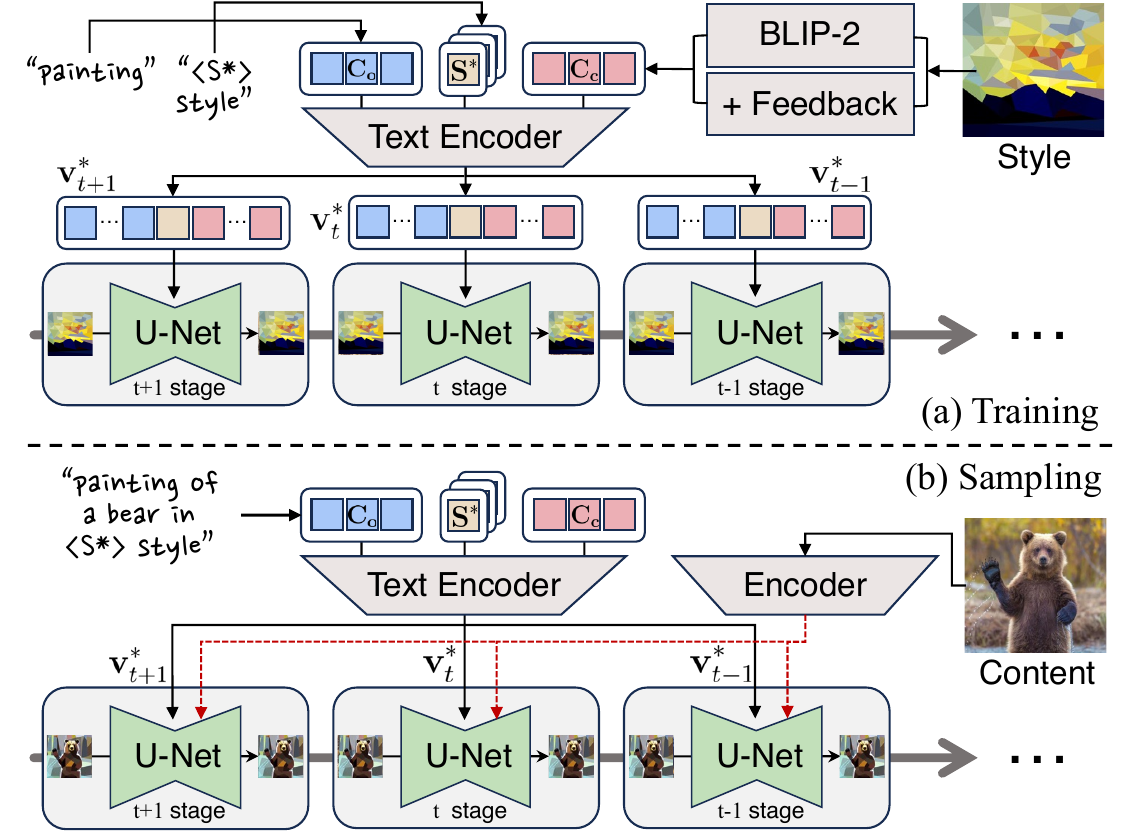}
\caption{\textbf{Model overview.} \textbf{(a)} \ours\ constructs training prompt with an opening text $C_o$, multi-stage style tokens $\mathbf{S^*}$, and a context description $C_c$, which is captioned with BLIP-2 and human feedback. \ours\ projects the training prompt into multi-stage textual embeddings $\mathbf{v}^* = \{v^*_1,\dots,v^*_{T}\}$, where $T$ is \# stages (a chunk of the denoising timestep). As a result, the denoising U-Net provides distinct textual information at each stage. \textbf{(b)} \ours\ prepares the textual embedding using a provided inference prompt. For style transfer, \ours\ employs ControlNet to comprehend the context information from a content image.}
\label{fig:model_overview}
\end{figure}
}

\newcommand{\figPrompt}{
\begin{figure*}[t]
\centering
\includegraphics[width=\linewidth]{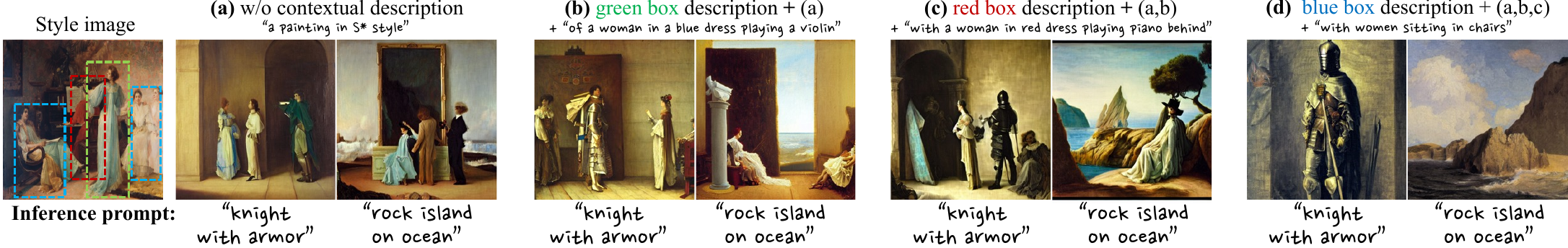}
\caption{\textbf{How does training prompt affect?} Given a style image, we construct training prompts with contextual descriptions (b$\sim$d). \textbf{(a)} Training without contextual description in the prompt; \textit{i.e.} trains the model with ``a painting in $S^*$ style".
The model tends to generate the images that contains objects and compositions from the style image (\textit{e.g.} standing and sitting audiences), instead of attributes depicted in the inference prompt.
\textbf{(b, c)} Training with partial contextual descriptions (the green and red boxes displayed in the style image, respectively). Such a tendency is significantly reduced, yet the model still synthesizes some objects from the style image (\textit{e.g.} sitting people in the blue box). \textbf{(d)} Training with full contextual descriptions. The model produces outputs that fully reflect the inference prompt without introducing any non-style attributes from the style image.}
\label{fig:prompt_preanalysis}
\end{figure*}
}

\newcommand{\figTIComp}{
\begin{figure*}[t]
\centering
\includegraphics[width=\linewidth]{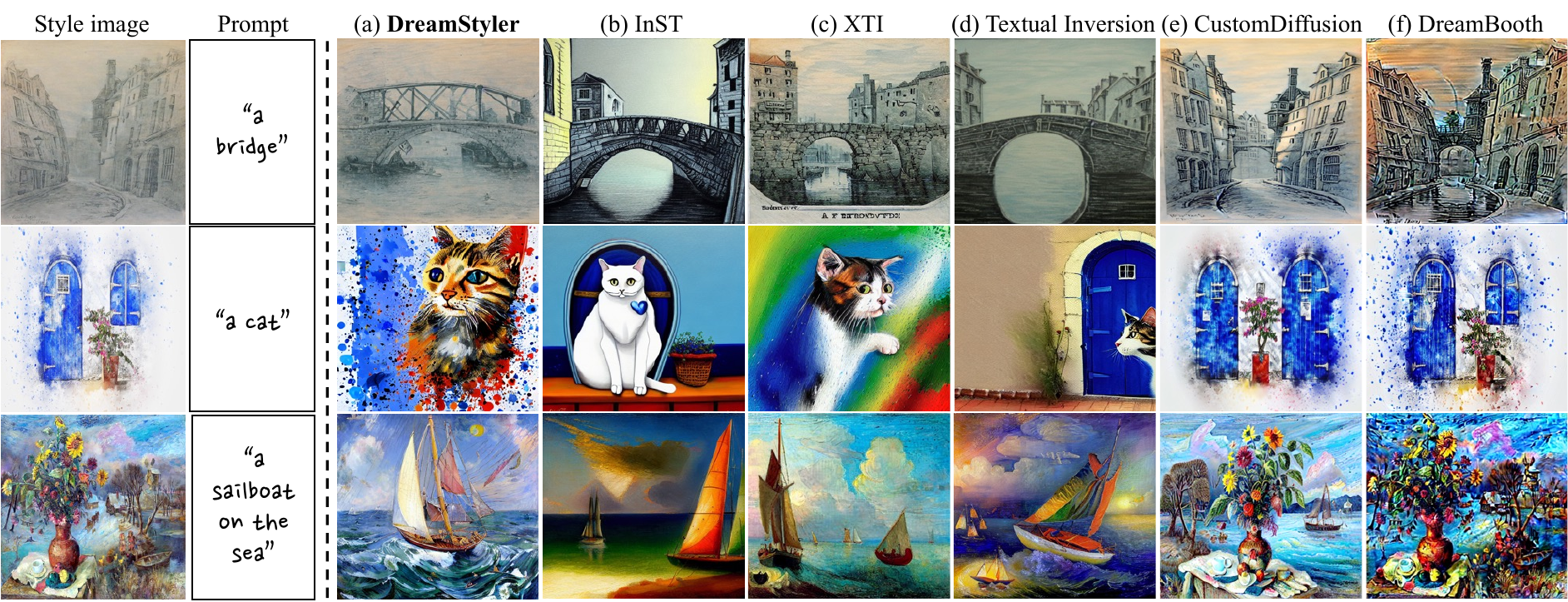}
\caption{\textbf{Qualitative comparison} on the style-guided text-to-image synthesis task.}
\label{fig:ti_comp}
\end{figure*}
}

\newcommand{\figTICompTradeoff}{
\begin{figure}[t]
\centering
\includegraphics[width=\linewidth]{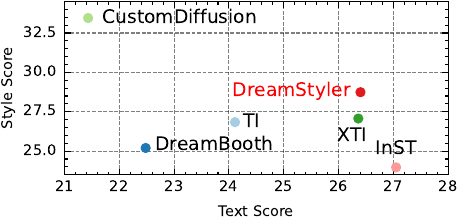}
\caption{\textbf{Performance of text and style scores} in style-guided text-to-image synthesis. \ours\ effectively balances these metrics and surpasses the majority of methods.}
\label{fig:ti_comp_tradeoff}
\end{figure}
}

\newcommand{\figSTComp}{
\begin{figure*}[t]
\centering
\includegraphics[width=\linewidth]{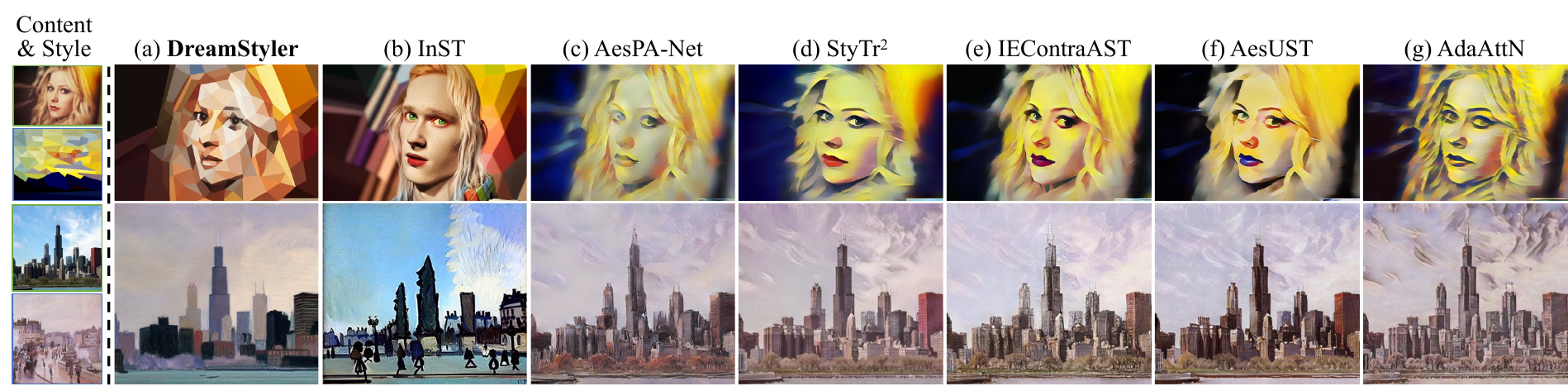}
\caption{\textbf{Qualitative comparison} on the style transfer task.}
\label{fig:st_comp}
\end{figure*}
}

\newcommand{\figSOComp}{
\begin{figure}[t]
\centering
\includegraphics[width=\linewidth]{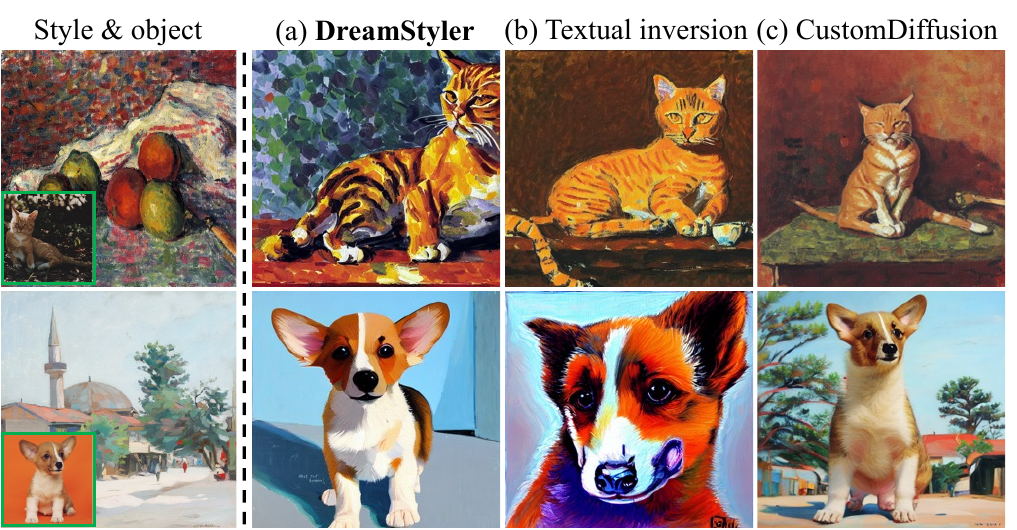}
\caption{\textbf{My object in my style.} Textual inversion faces challenges in accurately capturing both style and context from the reference images. Although CustomDiffusion successfully recreates the object's appearance, it tends to generate objects in a realistic style, which does not entirely match the target style image. On the other hand, \ours\ excels at synthesizing the object in the user-specified style.}
\label{fig:so_comp}
\end{figure}
}

\newcommand{\figMSTI}{
\begin{figure}[ht]
\centering
\includegraphics[width=\linewidth]{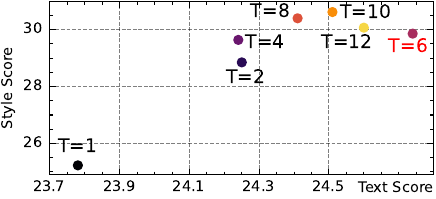}
\caption{\textbf{Study on the number of stages ($T$)} in multi-stage TI. We vary $T$ from 1 to 12 and select $T=6$ as the final model, considering the trade-off between text and style.}
\label{fig:ms_ti}
\end{figure}
}

\newcommand{\figStyleMixing}{
\begin{figure*}[ht]
\centering
\includegraphics[width=\linewidth]{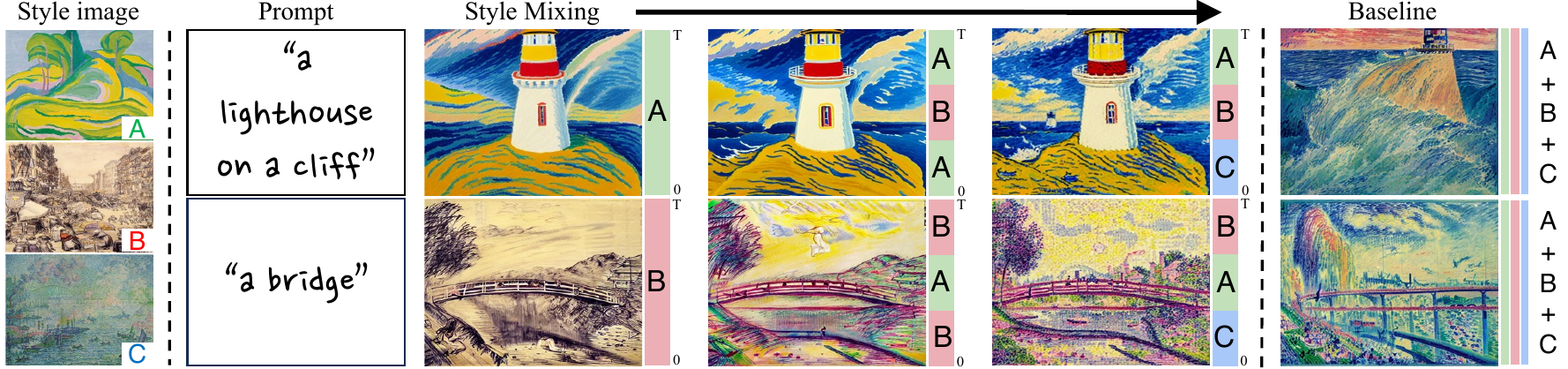}
\caption{\textbf{Style mixing.} Multi-stage TI facilitates style mixing from various style references. A user can customize a new style by substituting style tokens at different stages $t$. For example, the style token closer to $t=T$ tends to influence the structure of the image, while those closer to $t=0$ have a stronger effect on local and detailed attributes. For comparison, we display the baseline that employs all style tokens at every stage (\textit{i.e.} using ``A painting in $S^A_t$, $S^B_t$, $S^C_t$ style" at all stages).}
\label{fig:style_mixing}
\end{figure*}
}

\newcommand{\figMSAnalysis}{
\begin{figure}[ht]
\centering
\includegraphics[width=\linewidth]{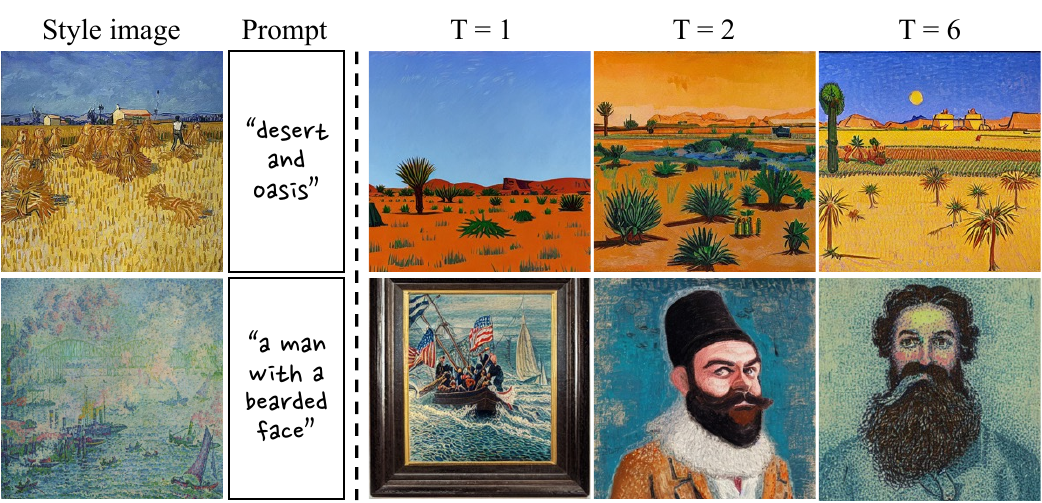}
\caption{\textbf{Visual comparison of varying $T$} in multi-stage TI. At $T=1$, the model fails in both style replication and prompt understanding. As $T$ increases, the style quality and text alignment are drastically enhanced.}
\label{fig:ms_analysis}
\end{figure}
}

\newcommand{\figPromptAnalysis}{
\begin{figure}[t]
\centering
\includegraphics[width=\linewidth]{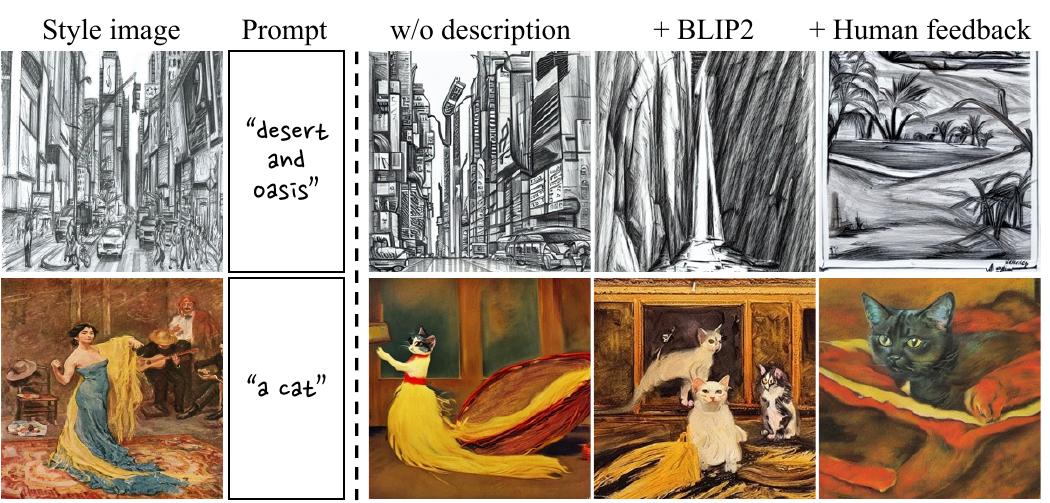}
\caption{\textbf{Comparison of three prompt strategies.} The model trained without contextual description struggles to disentangle style and context from the style image, generating elements present in the style reference (\textit{e.g.} the same composition in 1st row, a yellow dress in 2nd row). The contextual prompt alleviates this issue to some extent, but the BLIP2-based construction cannot completely eliminate it (\textit{e.g.} the same vanishing point in 1st row). The issue is thoroughly addressed when human feedback is utilized.}
\label{fig:prompt_analysis}
\end{figure}
}

\newcommand{\figGuidanceAnalysis}{
\begin{figure*}[ht]
\centering
\includegraphics[width=\linewidth]{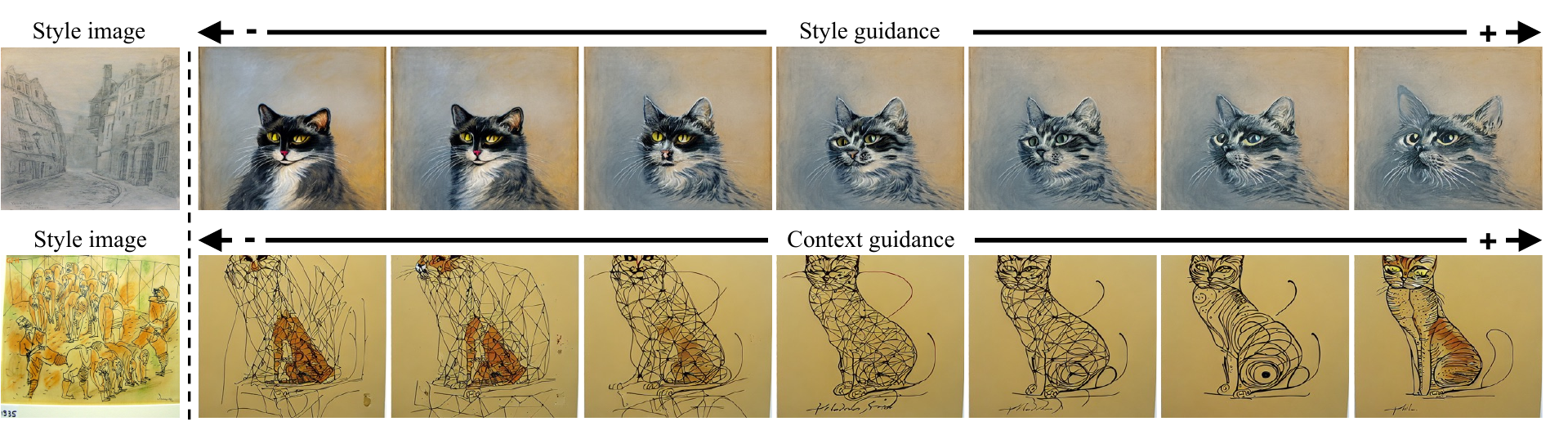}
\caption{\textbf{Study on the style and context guidance.} Inference prompt: ``A cat". By adjusting the scale parameters ($\lambda_s, \lambda_c$), we assess the influence of style and context guidance on the synthesized image. Increasing the style guidance strength causes the model to align more closely with the aesthetics of the style image; however, an excessive emphasis on style could compromise the context. Conversely, increasing the context guidance strength ensures the output corresponds with the inference prompt, but overly strong context guidance could deviate the output from the original style.}
\label{fig:guidance_analysis}
\end{figure*}
}


\newcommand{\figSupplTIComp}{
\begin{figure*}[p]
\centering
\includegraphics[width=\linewidth]{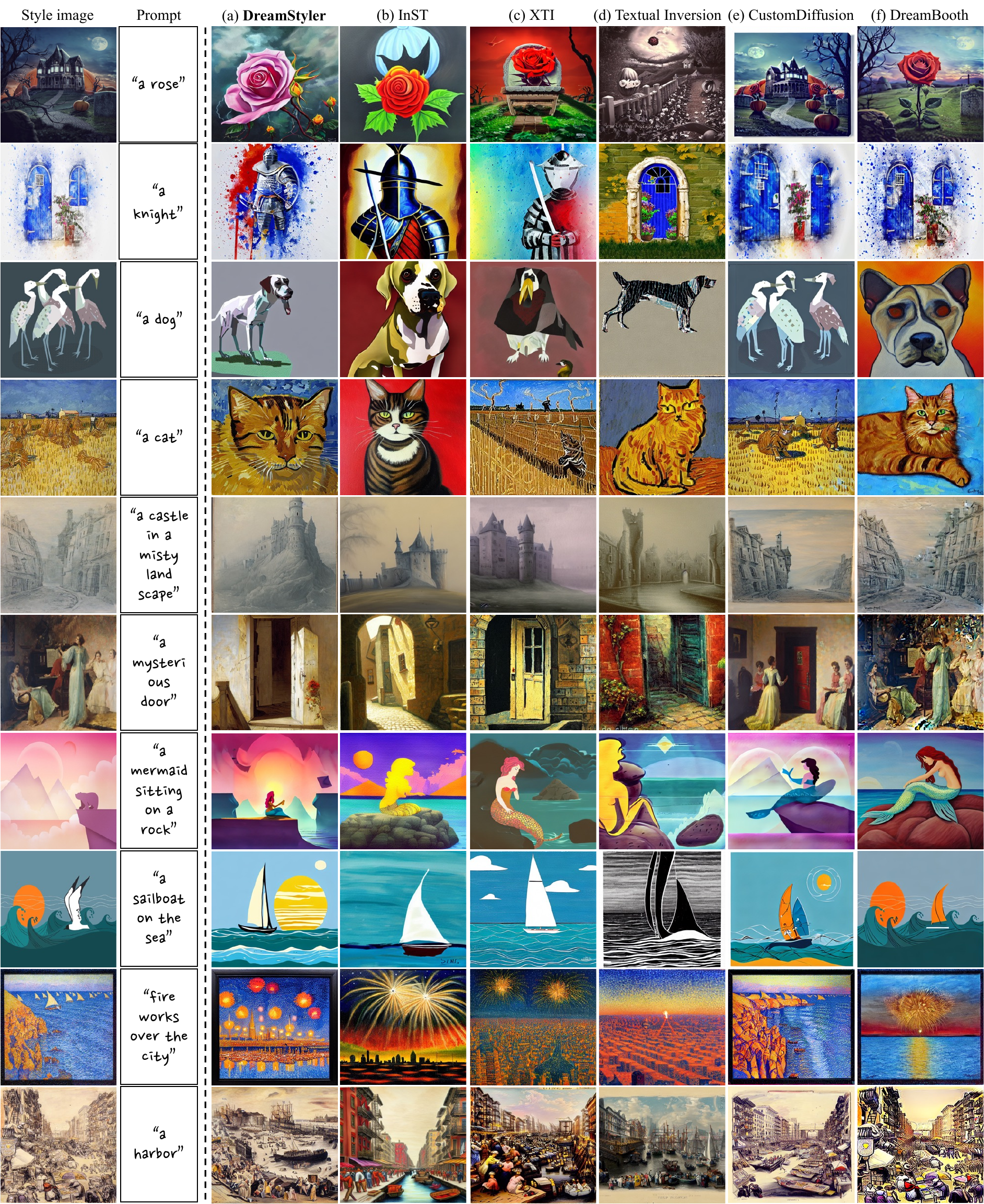}
\caption{\textbf{Additional qualitative comparison} on the style-guided text-to-image synthesis task.}
\label{fig:suppl_ti_comp}
\end{figure*}
}

\newcommand{\figSupplSTComp}{
\begin{figure*}[p]
\centering
\includegraphics[width=\linewidth]{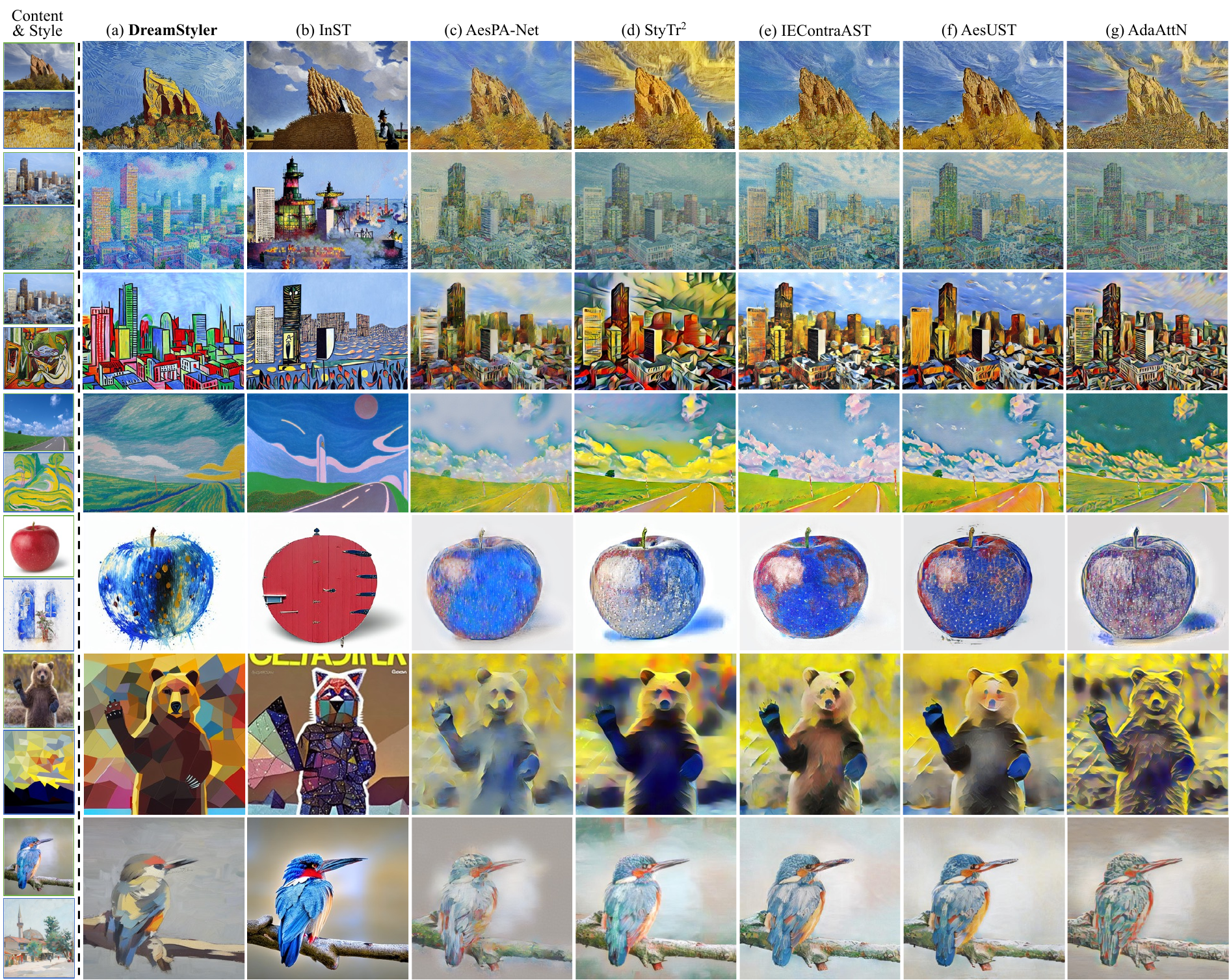}
\caption{\textbf{Additional qualitative comparison} on the style transfer task.}
\label{fig:suppl_st_comp}
\end{figure*}
}

\newcommand{\figSupplFewshot}{
\begin{figure}[t]
\centering
\includegraphics[width=\linewidth]{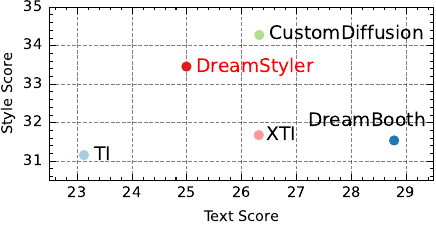}
\caption{\textbf{Performance of few-shot style-guided} text-to-image synthesis. All methods are trained using five artistic style images. In a few-shot training regime, the model optimization-based approaches (DreamBooth, CustomDiffusion) significantly enhance text score. This is because they now can distinguish context from style by referring to multiple style images, leveraging their powerful capacity inherited from learning the model parameters. However, previous textual inversion-based methods (TI, XTI) cannot enjoy using multiple images, showing lower style and text scores. \ours\ strikes a good balance, exhibiting significant improvement over other textual inversion-based methods.}
\label{fig:suppl_fewshot}
\end{figure}
}

\newcommand{\figSupplAblGuidance}{
\begin{figure}[t]
\centering
\includegraphics[width=\linewidth]{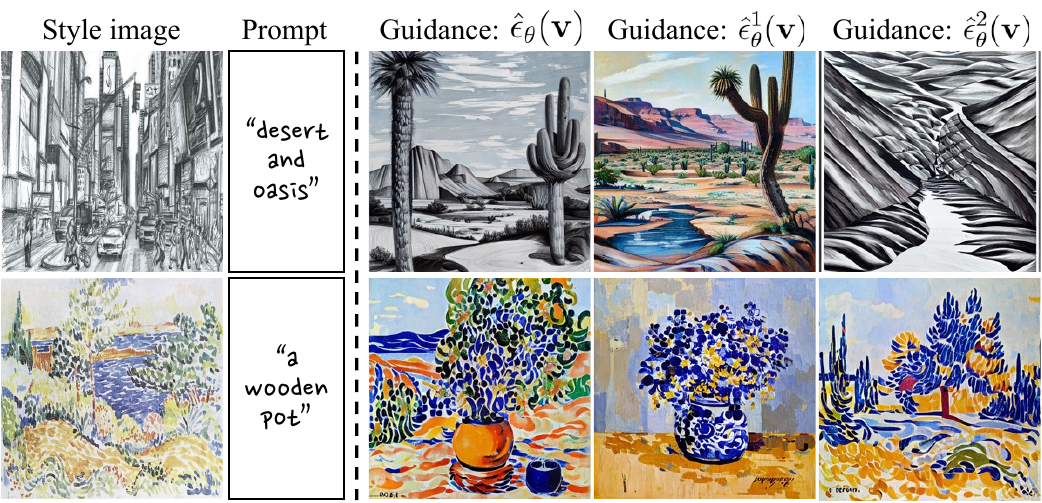}
\caption{\textbf{Comparison of different guidance forms.} Applying $\hat{\epsilon}_\theta^1(\mathbf{v})$ guidance effectively captures the context of a given inference prompt. However, it sometimes fails to adopt the style of a reference image. Conversely, $\hat{\epsilon}_\theta^2(\mathbf{v})$ guidance adeptly aligns the stylistic elements with the reference image but often overlooks certain contexts in the inference prompt. By merging these guidance terms, as depicted in \eref{eq:guidance} in the main text, the model achieves a well-balanced expression of both style and context.}
\label{fig:suppl_abl_guidance}
\end{figure}
}

\newcommand{\figSupplAblStyleTransfer}{
\begin{figure}[t]
\centering
\includegraphics[width=\linewidth]{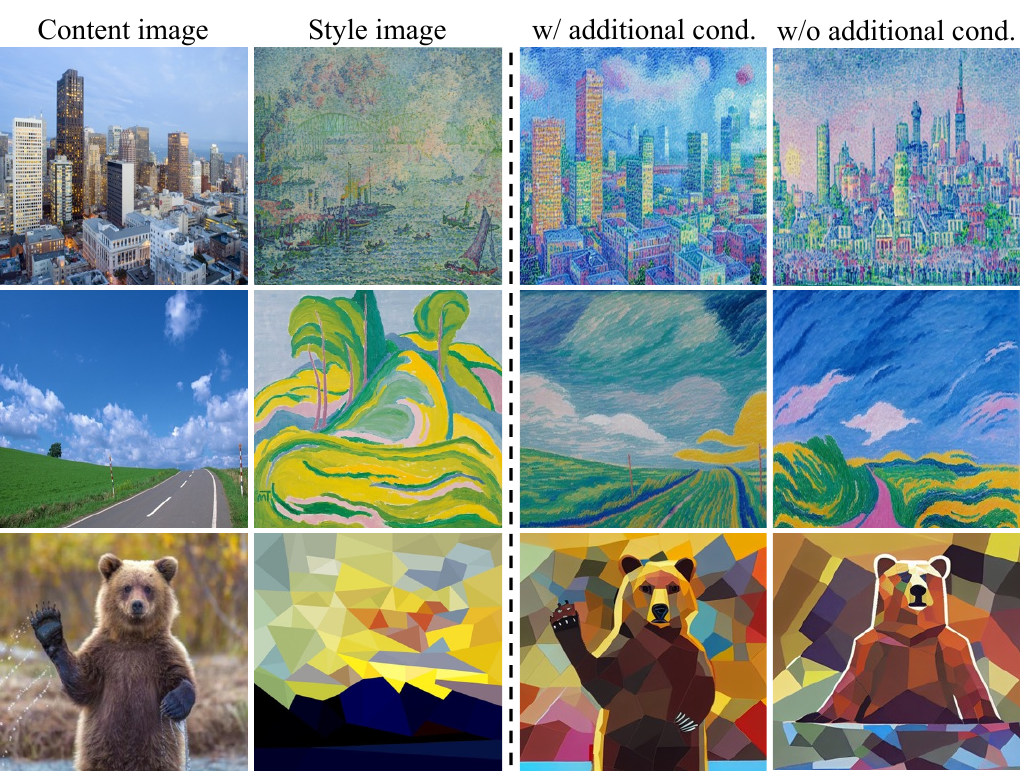}
\caption{\textbf{Study on the role of additional conditions in style transfer.} When using additional conditions encoded through ControlNet~\cite{zhang2023adding}, the outputs faithfully reflect the structure of content images. In contrast, outputs without additional condition and relying solely on image inversion~\cite{meng2021sdedit} considerably changes the structure. This issue is also observed in other inversion-based methods~\cite{zhang2023inversion,ahn2023interactive}.}
\label{fig:suppl_abl_style_transfer}
\end{figure}
}

\newcommand{\figSupplSG}{
\begin{figure*}[p]
\centering
\begin{subfigure}[b]{1.0\linewidth}
\centering
\includegraphics[width=\linewidth]{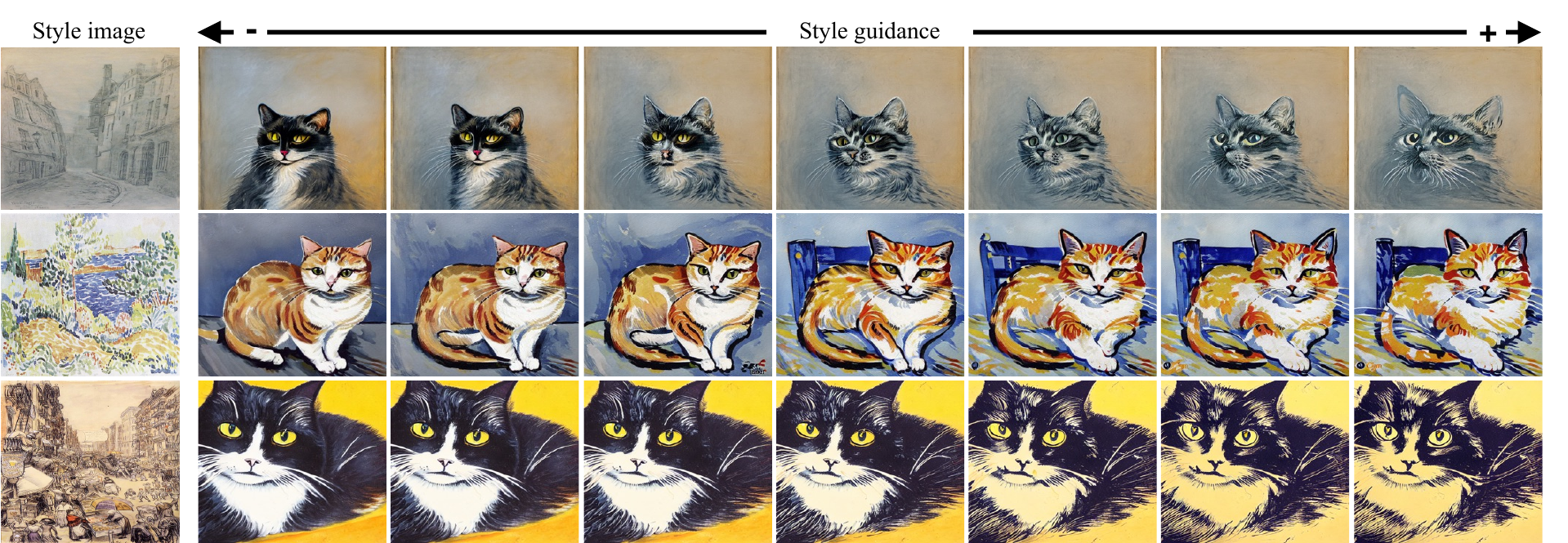}
\caption{Significant effects of the style guidance.}
\end{subfigure}
\\
\begin{subfigure}[b]{1.0\linewidth}
\centering
\includegraphics[width=\linewidth]{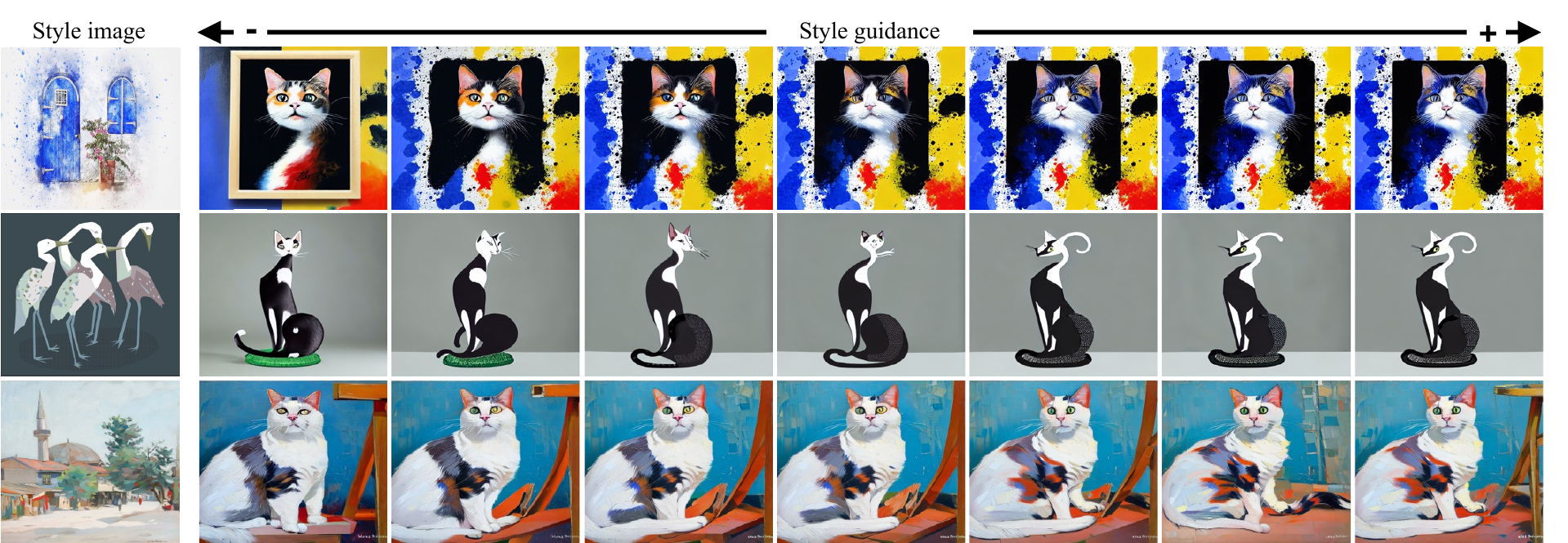}
\caption{Moderate effects of the style guidance.}
\end{subfigure}
\caption{\textbf{When is the style guidance beneficial?} Inference prompt: ``A cat". We investigate the role of the style guidance in stylistic representation. \textbf{(a)} Our findings show that the style guidance significantly impacts style expression when generated samples with zero-guidance (leftmost ones) fails to adequately capture the style image. \textbf{(b)} In contrast, when the samples without the style guidance effectively embody stylistic nuances, the introduction of the style guidance has minimal effect. We hypothesize that the style guidance is particularly beneficial for images with high pattern complexity (as shown in (a)), so that the model cannot easily adapt to. In cases of low pattern complexity (as shown in (b)), its influence appears to be marginal.}
\label{fig:suppl_sg}
\end{figure*}
}

\newcommand{\figSupplCG}{
\begin{figure*}[p]
\centering
\begin{subfigure}[b]{1.0\linewidth}
\centering
\includegraphics[width=\linewidth]{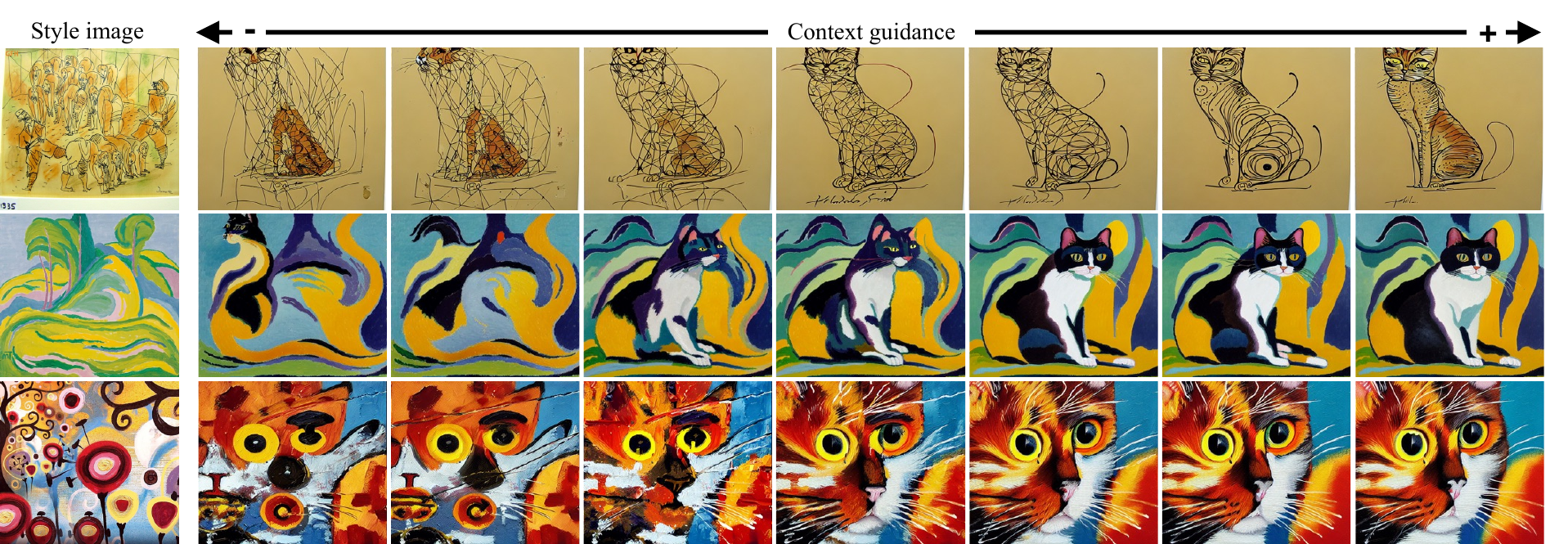}
\caption{Significant effects of the context guidance.}
\end{subfigure}
\\
\begin{subfigure}[b]{1.0\linewidth}
\centering
\includegraphics[width=\linewidth]{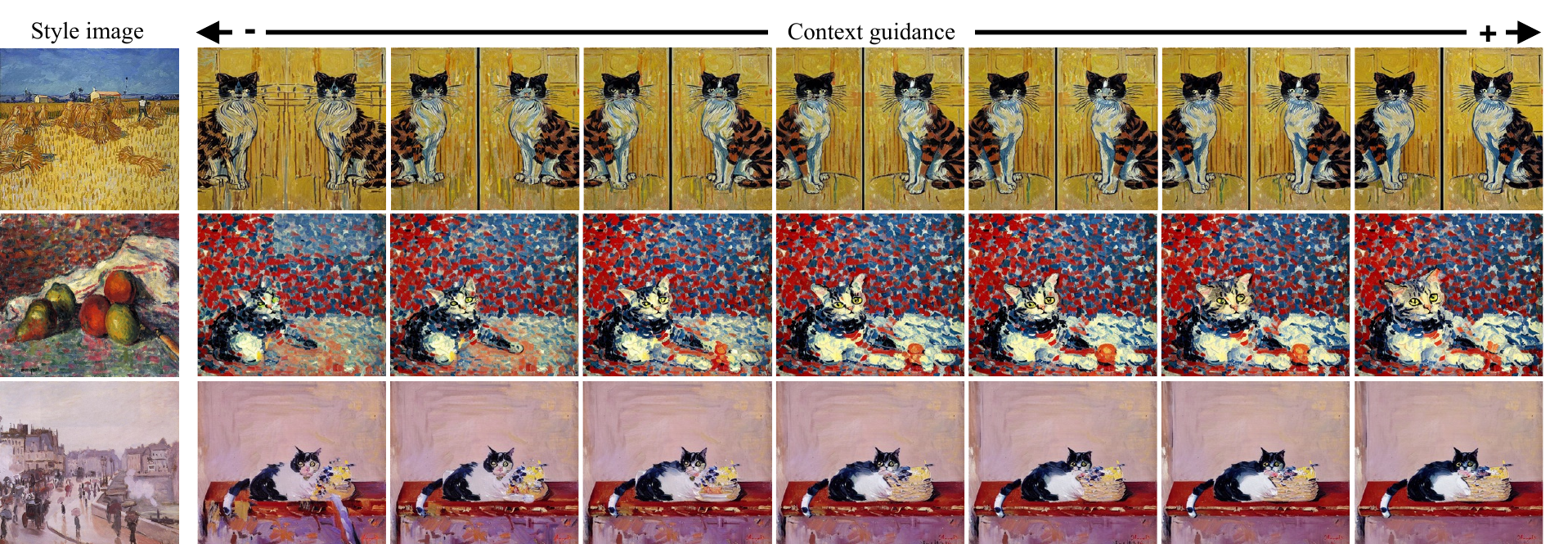}
\caption{Moderate effects of the context guidance.}
\end{subfigure}
\caption{\textbf{When is the context guidance beneficial?} Inference prompt: ``A cat".  We investigate the role of the context guidance in the context (subject) production aspect. \textbf{(a)} Our findings show that the context guidance substantially impacts to appearance of subjects when generated samples with zero-guidance (leftmost ones) exhibit abstract subject structures. \textbf{(b)} In contrast, when subjects already appear detailed even without the context guidance, the introduction of the context guidance has minimal effect. We hypothesize that the context guidance is particularly beneficial for images with a high degree of structural abstraction (as shown in (a)). In contrast, for the images that present detailed textures, its influence appears to be marginal.}
\label{fig:suppl_cg}
\end{figure*}
}

\newcommand{\figSupplTrainingA}{
\begin{figure*}[p]
\centering
\includegraphics[width=0.7\linewidth]{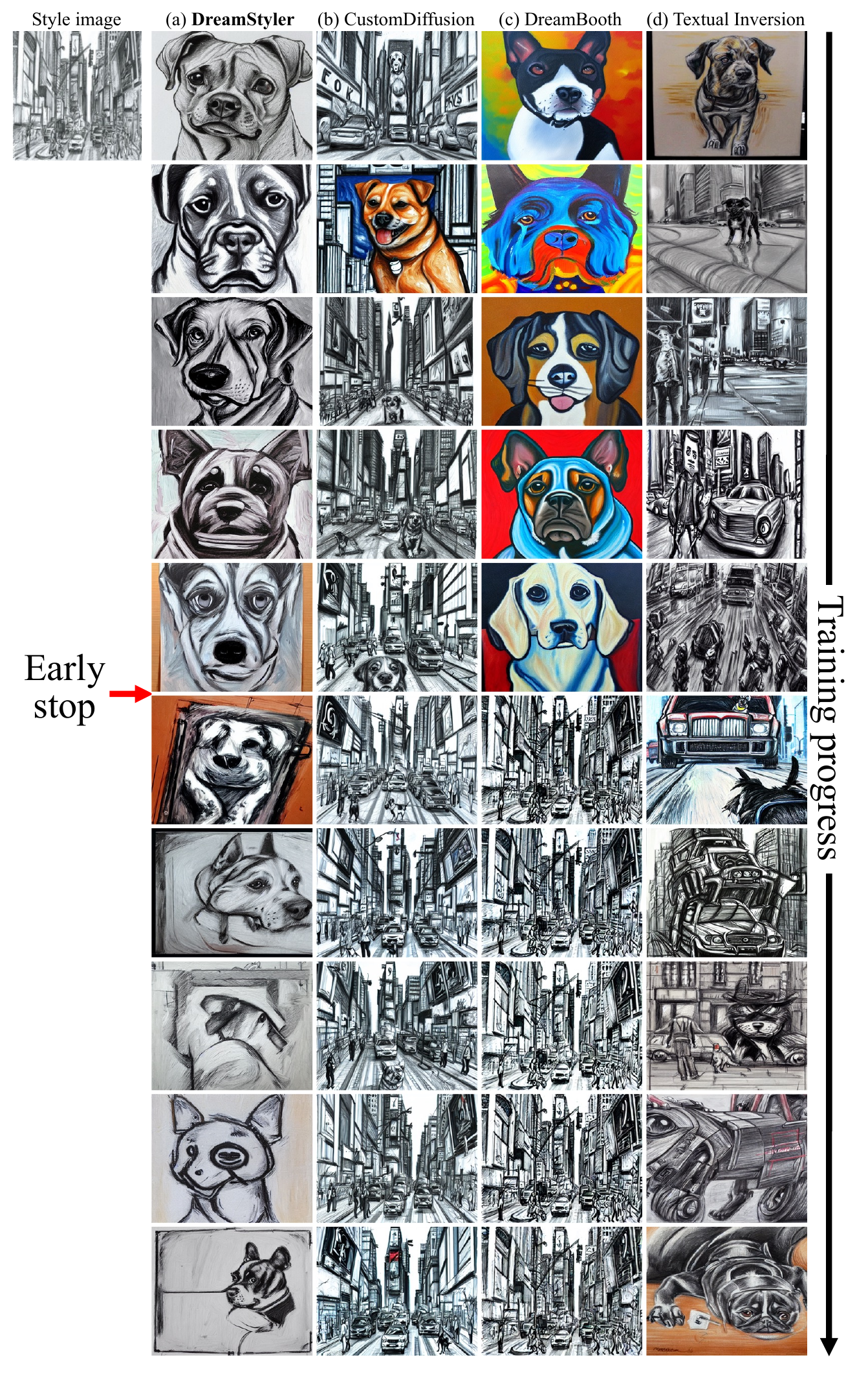}
\caption{\textbf{Comparison of training progress.} Inference prompt: ``A painting of a dog". We double the training steps for each model and visualize their training tendencies. ``Early stop" refers to the original (official) training step which we also used in the paper. Model optimization-based methods tend to suffer from overfitting to the style image, and we observed that the occurrence of overfitting varies depending on the style image. Notably, DreamBooth often fails to reflect the style image initially and then abruptly overfits to it. Since it's challenging to predict when overfitting will begin, users would need to monitor all intermediate training samples, which is impractical for production-level tasks. While textual inversion-based methods (including ours) are immune to overfitting, the vanilla TI often fails to accurately capture both the style images and the inference prompt. In contrast, \ours\ avoids overfitting and consistently demonstrates high-quality stylization in terms of both style and context.}
\label{fig:suppl_training_A}
\end{figure*}
}

\newcommand{\figSupplTrainingB}{
\begin{figure*}[p]
\centering
\includegraphics[width=0.7\linewidth]{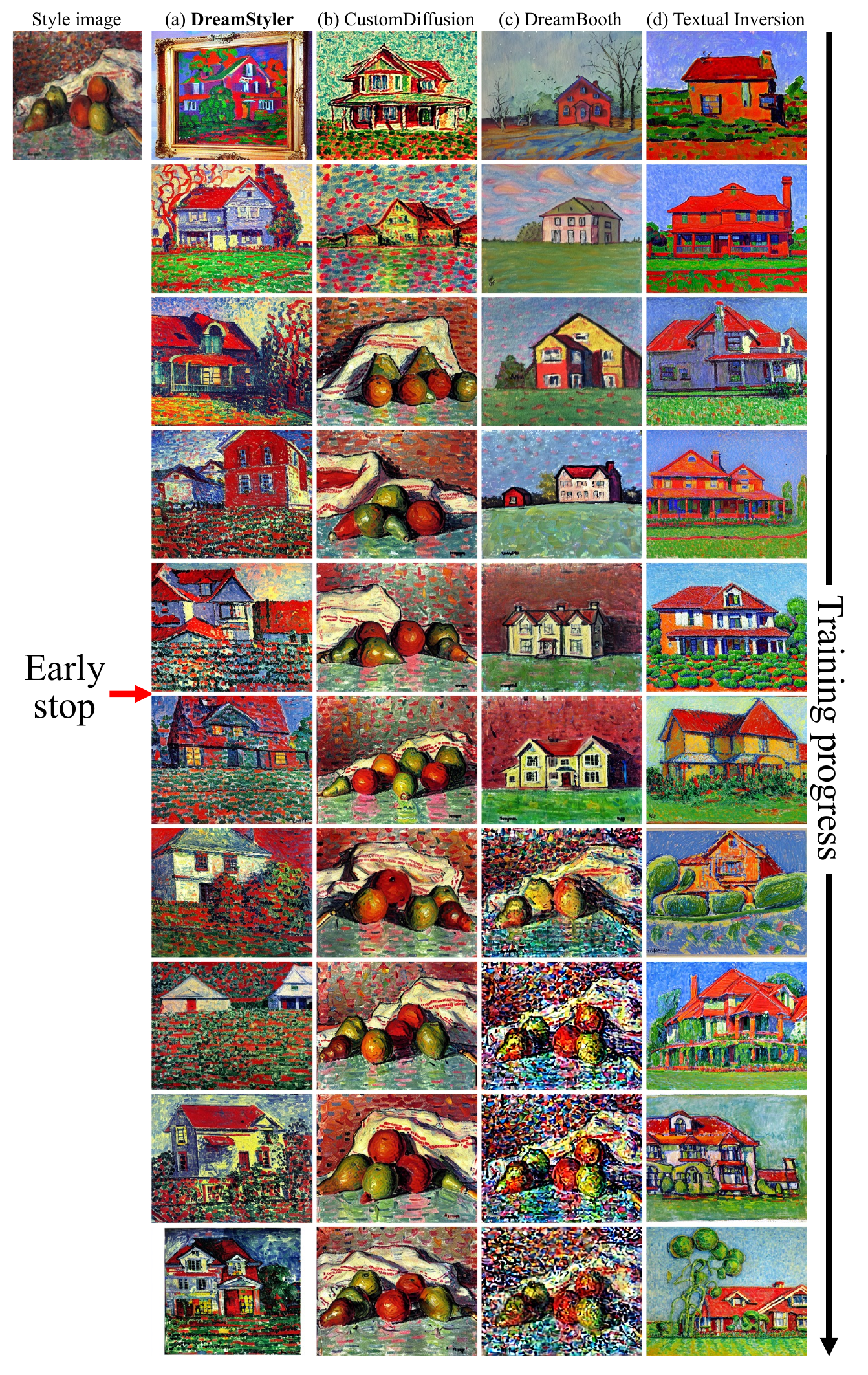}
\caption{\textbf{Comparison of training progress.} Inference prompt: ``A painting of a house".}
\label{fig:suppl_training_B}
\end{figure*}
}

\newcommand{\figSupplTrainingC}{
\begin{figure*}[p]
\centering
\includegraphics[width=0.7\linewidth]{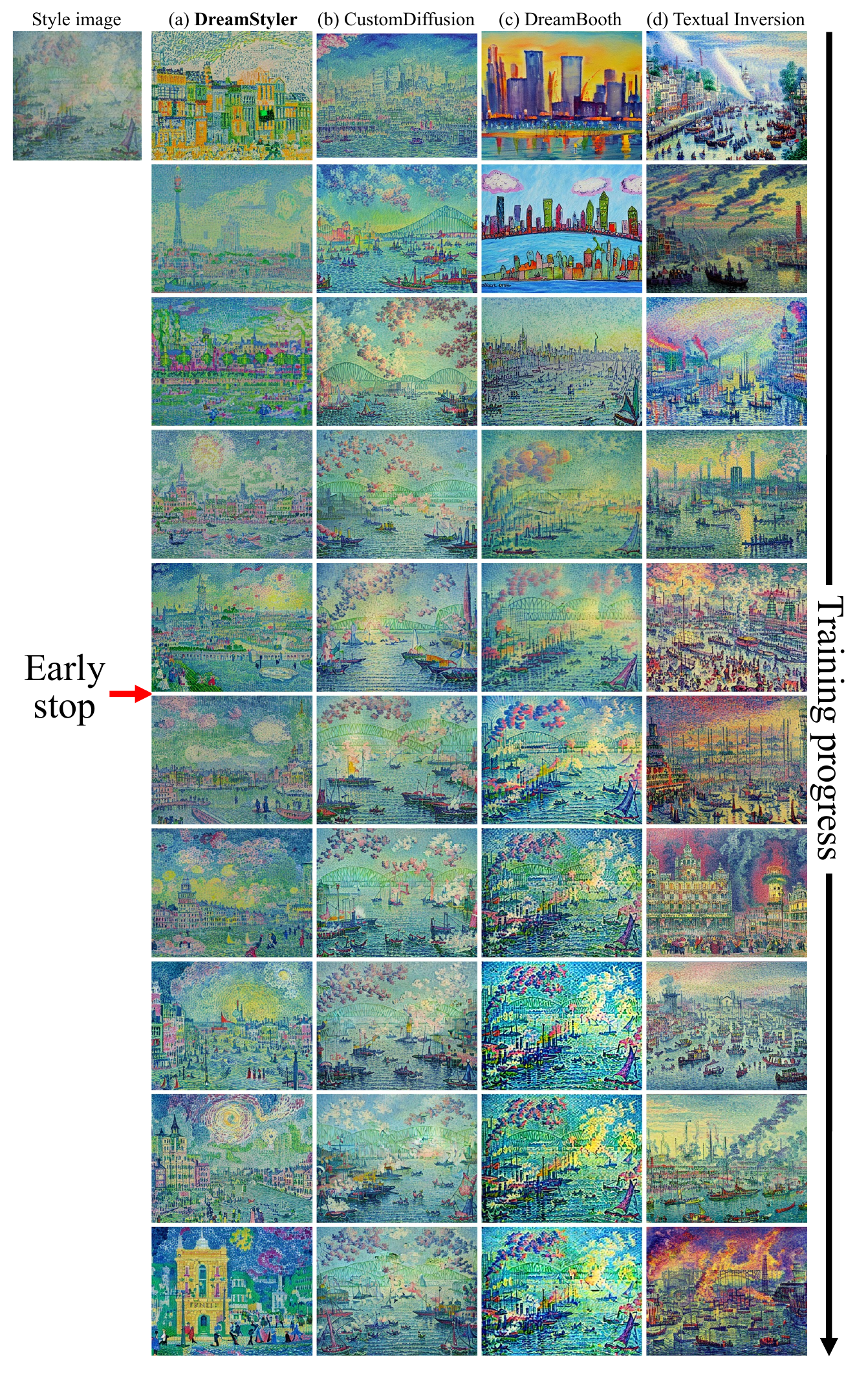}
\caption{\textbf{Comparison of training progress.} Inference prompt: A painting of a city".}
\label{fig:suppl_training_C}
\end{figure*}
}
\newcommand{\tableTIComp}{
\begin{table}[t]
\centering
\setlength\tabcolsep{4.5pt}
\small
\begin{tabular}{l | c c c}
\hline
\multirow{2}{*}{Method} & Text  & Style & User \\
& Score & Score & Score \\
\hline\hline
Textual Inversion~\cite{gal2022image}	& 24.11	& 26.84 & 2.1\% \\
DreamBooth~\cite{ruiz2023dreambooth}	& 22.48	& 25.20 & 3.9\% \\
CustomDiffusion~\cite{kumari2023multi}	& 21.43	& \textbf{33.45} & \underline{4.8}\% \\
XTI~\cite{voynov2023p+} & 26.36	& 27.07 & 4.5\%\\
InST~\cite{zhang2023inversion} &	\textbf{27.05} & 23.97 & 1.8\%\\
\hline
\textbf{\ours  \ (Ours)} & \underline{26.40} & \underline{28.74} & \textbf{82.9\%}\\
\hline
\end{tabular}
\caption{\textbf{Quantitative comparison} on the style-guided text-to-image synthesis task. \textbf{Bold}: best, \underline{underline}: second best.}
\label{table:ti_comp}
\end{table}
}

\newcommand{\tableSTComp}{
\begin{table}[t]
\centering
\small
\begin{tabular}{l | c c c}
\hline
\multirow{2}{*}{Method} & Text & Image & User \\
& Score & Score & Score \\
\hline\hline
AdaAttN~\cite{liu2021adaattn} & 56.67 & 56.76 & 8.6\% \\
AesUST~\cite{wang2022aesust} & 58.05 & 58.09 & 6.8\%\\
IEContraAST~\cite{chen2021artistic} & 59.38 & 59.42 & 8.6\%\\
$\textnormal{StyTr}^2$~\cite{deng2022stytr2} & 56.18 & 56.28 & \underline{21.2\%}\\
AesPA-Net~\cite{hong2023aespanet} & 58.08 & 58.15 & 8.6\%\\
InST~\cite{zhang2023inversion} & \underline{65.32} & \underline{65.37} & 2.3\%\\
\hline
\textbf{\ours  \ (Ours)} & \textbf{66.04} & \textbf{66.05} & \textbf{44.1\%} \\
\hline
\end{tabular}
\caption{\textbf{Quantitative comparison} on the style transfer task.}
\label{table:st_comp}
\end{table}
}

\newcommand{\tableAblation}{
\begin{table}[t]
\centering
\begin{tabular}{l | c c}
\hline
Method & Text Score & Style Score \\
\hline\hline
Baseline~\cite{gal2022image} & 23.78 & 25.23 \\
+ Multi-Stage TI & 24.74 & \textbf{29.86} \\
+ Context-Aware Prompt & 24.65 & 29.50 \\
+ S\&C Guidance (\textbf{Ours}) & \textbf{25.38} & 29.62 \\
\hline
\end{tabular}
\caption{\textbf{Model ablation study.} Upon the textual inversion baseline~\cite{gal2022image}, we attach the proposed components to measure the effectiveness of our method.}
\label{table:ablation}
\end{table}
}


\newcommand{\tableSupplSetting}{
\begin{table}[ht]
\centering
\begin{tabular}{c|lc}
\hline
Phase & Hyperparameter & Value \\
\hline\hline
\multirow{4}{*}{Optimization} & Optimization steps & 500 \\
& Learning rate & 0.002 \\
& Batch size & 8 \\
& $T$ (multi-stage TI) & 6 \\
\hline
\multirow{4}{*}{Inference} & Inference steps & 25 \\
& Scheduler & DPM \\
& $\lambda_n$ (null guidance) & 5.0 \\
& $\lambda_s$ (style guidance) & 0.5$\sim$3.0 \\
& $\lambda_c$ (context guidance) & 3.0 \\
\hline
\end{tabular}
\caption{\textbf{Detailed hyperparameters of \ours.} }
\label{table:suppl_settings}
\end{table}
}

\newcommand{\tableExPrompts}{
\begin{table*}[t]
\centering
\begin{tabular}{l}
\hline
Evaluation prompt \\
\hline\hline
A painting of a house in the style of $S^*$ \\
A painting of a dog in the style of $S^*$ \\
A painting of a robot in the style of $S^*$ \\
A painting of a crying woman in the style of $S^*$ \\
A painting of a wooden pot in the style of $S^*$ \\
A painting of an ocean on a cloudy day in the style of $S^*$ \\
A painting of a city in the style of $S^*$ \\
A painting of a full moon in the mountains in the style of $S^*$ \\
A painting of a temple in the forest in the style of $S^*$ \\
A painting of a cat in the style of $S^*$ \\
A painting of a rose in the style of $S^*$ \\
A painting of a man with a bearded face in the style of $S^*$ \\
A painting of a solar system in the style of $S^*$ \\
A painting of a spaceship in the style of $S^*$ \\
A painting of a highway in the style of $S^*$ \\
A painting of a bridge in the style of $S^*$ \\
A painting of an apple and a banana in the style of $S^*$ \\
A painting of fireworks over the city in the style of $S^*$ \\
A painting of a harbor in the style of $S^*$ \\
A painting of a rock island and an ocean in the style of $S^*$ \\
A painting of a blossoming cherry tree in the style of $S^*$ \\
A painting of a snowy mountain peak in the style of $S^*$ \\
A painting of a knight in the style of $S^*$ \\
A painting of a marketplace in the style of $S^*$ \\
A painting of a desert and an oasis in the style of $S^*$ \\
A painting of elephants at sunset in the style of $S^*$ \\
A painting of a crowd under the stars in the style of $S^*$ \\
A painting of a lighthouse on a cliff in the style of $S^*$ \\
A painting of a mermaid sitting on a rock in the style of $S^*$ \\
A painting of a haunted house on a hill in the style of $S^*$ \\
A painting of a cafe in the morning in the style of $S^*$ \\
A painting of a castle in a misty landscape in the style of $S^*$ \\
A painting of birds taking flight in the style of $S^*$ \\
A painting of a family in the style of $S^*$ \\
A painting of a waterfall in a forest in the style of $S^*$ \\
A painting of a night of shooting stars in the style of $S^*$ \\
A painting of a sailboat on the sea in the style of $S^*$ \\
A painting of a girl under a tree in the style of $S^*$ \\
A painting of a fisherman casting his net at sunrise in the style of $S^*$ \\
\hline
\end{tabular}
\caption{\textbf{Inference prompts used in the model evaluation.} }
\label{table:suppl_ex_prompts}
\end{table*}
}

\newcommand{\tableUserStudy}{
\begin{table*}[t]
\centering
\begin{tabular}{c|l}
\hline
Task & Instruction and Question \\
\hline\hline
Text-to-Image & \makecell[l]{\textbf{Instruction:}\\
Each question provides 1) a style image, 2) a text prompt, and 3) 6 generated samples.\\
Samples are results of synthesizing context of the text prompt to the style depicted in the style image.\\
Please select the sample that best represents both the text prompt and the style of the style image.\\\\
\textbf{Question:}\\
Which sample best captures the style of the style image and the content of the text?
}
\\\hline
Style Transfer & \makecell[l]{\textbf{Instruction:}\\
Each question provides 1) a style image, 2) a content image, and 3) 7 generated samples.\\
Samples are results of adapting the content image to the style depicted in the style image.\\
Please select the sample that best resembles with the artistic style of the artists' style image,\\as if that artist used the content image as a reference in their painting.\\\\
\textbf{Question:}\\
If the artist of the style image had used the content image as a reference,\\which sample best embodies their artistic style?}

\\
\hline
\end{tabular}
\caption{\textbf{Questionnaires used in the user study.} }
\label{table:suppl_user_study}
\end{table*}
}

\begin{abstract}
Recent progresses in large-scale text-to-image models have yielded remarkable accomplishments, finding various applications in art domain.
However, expressing unique characteristics of an artwork (\textit{e.g.} brushwork, colortone, or composition) with text prompts alone may encounter limitations due to the inherent constraints of verbal description.
To this end, we introduce \ours, a novel framework designed for artistic image synthesis, proficient in both text-to-image synthesis and style transfer.
\ours\ optimizes a multi-stage textual embedding with a context-aware text prompt, resulting in prominent image quality.
In addition, with content and style guidance, \ours\ exhibits flexibility to accommodate a range of style references.
Experimental results demonstrate its superior performance across multiple scenarios, suggesting its promising potential in artistic product creation. Project page: \url{https://nmhkahn.github.io/dreamstyler/}.
\end{abstract}
\section{Introduction}

\begin{quote}
``\textit{Painting is silent poetry.}" --- Simonides, Greek poet
\end{quote}

\figTeaser{}

Recent text-to-image models have shown unprecedented proficiency in translating natural language into compelling visual imagery~\cite{saharia2022photorealistic,ramesh2022hierarchical,rombach2022high}.
These have emerged in the realm of art, providing inspiration and even assisting in crafting tangible art pieces.
In the AI-assisted art production workflow, artists typically utilize various descriptive prompts that depict the style and context to generate their desired image.
However, the unique styles of a painting, its intricate brushwork, light, colortone, or composition, cannot be easily described in a single word.
For instance, dare we simplify the entirety of Vincent Van Gogh's lifelong artworks as just one word, `Gogh style'?
Text descriptions cannot fully evoke his unique style in our imagination --- his vibrant color, dramatic light, and rough yet vigorous brushwork.

Beyond text description, recent studies~\cite{gal2022image, ruiz2023dreambooth} embed specific attributes of input images into latent space.
While they effectively encapsulate a novel \textit{object}, we observed that they struggle to personalize \textit{style} of a painting.
For instance, model optimization-based methods~\cite{ruiz2023dreambooth, kumari2023multi} are highly susceptible to overfitting and often neglect inference prompts, which is not ideal for real-world production (please refer to the Suppl. for more details).
Textual inversion-based methods~\cite{gal2022image,voynov2023p+}, in contrast, effectively reflect the inference prompt but fail to replicate style, possibly due to the limited capacity of the learned embeddings.
This is because capturing style, from global elements (\textit{e.g.} colortone) to local details (\textit{e.g.} detailed texture), is challenging when relying solely on a single embedding token.

In this work, we present \textbf{\ours}, a novel single (one-shot) reference-guided artistic image synthesis framework designed for the text-to-image generation and style transfer tasks (\fref{fig:teaser}).
We encapsulate the intricate styles of artworks into CLIP text space.
\ours\ is grounded in textual inversion (TI), chosen for the inherent flexibility that stems from its prompt-based configuration.
To overcome the limitations of TI, we introduce an extended textual embedding space, $\mathcal{S}$ by expanding textual embedding into the denoising timestep domain (\fref{fig:model_overview}).
Based on this space, we propose a multi-stage TI, which maps the textual information into the $\mathcal{S}$ space.
It accomplishes by segmenting the entire diffusion process into multiple \textit{stages} (a chunk of timesteps) and allocating each textual embedding vector to the corresponding stage.
The exploitation of the timestep domain in textual inversion significantly improves the overall efficacy of artistic image synthesis.
This enhancement stems from the increased capacity of the personalized module, as well as the utilization of prior knowledge suggesting that different denoising diffusion steps contribute differently to image synthesis~\cite{balaji2022ediffi,choi2022perception}.

We further propose a context-aware prompt augmentation that simply yet proficiently decouples the style and context information from the reference image.
With our approach, the personalization module can embed style features solely into its textual embeddings, ensuring a more faithful reflection of the reference's style.
To further refine the artistic image synthesis, we introduce a style and context guidance, inspired by classifier-free guidance~\cite{ho2022classifier}.
Our guidance bisects the guidance term into style and context components, enabling individual control.
Such a guidance design allows users to tailor the outputs based on their preferences or intricacy of the reference image's style.

We validate the effectiveness of \ours\ through a broad range of experiments.
\ours\ not only demonstrates advanced artistic image synthesis but also paves the new way of applying text-to-image diffusion models to the realms of artistic image synthesis and style transfer tasks.

\section{Related Work}
\noindent\textbf{Personalized text-to-image synthesis.}
Since the latent-based text conditional generation has been explored~\cite{rombach2022high}, following studies~\cite{saharia2022photorealistic,ramesh2022hierarchical,li2022upainting} have further contributed to enhancing text-to-image synthesis with CLIP~\cite{radford2021learning} guidance.
Furthermore, Textual inversion~\cite{gal2022image}, DreamBooth~\cite{ruiz2023dreambooth} and CustomDiffusion~\cite{kumari2023multi} introduced approaches that leverage 3-5 images of the subject to personalize semantic features.
Recently, \citet{voynov2023p+} proposed $\mathcal{P+}$ space, which consists of multiple textual conditions, derived from per-layer prompts.
Although they showed promising results in penalization of diffusion models, there are still limitations to fully capturing precise artistic style representations.
In contrast, \ours\ considers the denoising timestep to accommodate temporal dynamics in the diffusion process, achieving high-quality artistic image generation.

\smallskip
\noindent\textbf{Paint by style.}
Neural style transfer renders the context of a source with a style image.
Since \citet{gatys2016image}, studies have been devoted to enhancing the transfer networks for more accurate and convincing style transfer. 
Notably, AdaIN~\cite{huang2017arbitrary} and AdaAttN~\cite{liu2021adaattn} investigated matching the second-order statistics of content and style images.
AesPA-Net~\cite{hong2023aespanet} and StyTr$^2$~\cite{deng2022stytr2} adopted recent architectures such as attention or transformer for high-fidelity neural style transfer.
Recently, InST~\cite{zhang2023inversion} utilized the diffusion models by introducing the image encoder to inverse style images into CLIP spaces.

\figModelOverview{}
\figPrompt{}

\section{Method}
\noindent\textbf{Preliminary: Stable Diffusion (SD).}
\ours\ is built upon SD~\cite{rombach2022high}.
SD projects an input image $x$ into a latent code, $z = E(x)$ using an encoder $E$, while decoder $D$ transforms the latent code back into pixel space, \textit{i.e.} $x' = D(z')$.
The diffusion model creates a new latent code $z'$ by conditioning on additional inputs such as a text prompt $y$. The training objective of SD is defined as:
\begin{equation}
\mathcal{L} = \mathbb{E}_{z\sim E(x), y, \epsilon\sim N(0, 1), t} [||\epsilon - \epsilon_{\theta}(z_t, t, c(y))||^2_2].
\label{eq:sd}
\end{equation}

At each timestep $t$, the denoising network $\epsilon_{\theta}$ reconstructs the noised latent code $z_t$, given the timestep $t$ and a conditioning vector $c(y)$. To generate $c(y)$, each token from a prompt is converted into an embedding vector, which is then passed to the CLIP text encoder~\cite{radford2021learning}.

\smallskip
\noindent\textbf{Preliminary: Textual Inversion (TI).} \citet{gal2022image} proposed a method to personalize a pre-trained text-to-image model by incorporating a novel embedding representing the intended concept.
To personalize the concept, they initialize a word token $S^*$ and its corresponding vector $v^*$, situated in the textual conditioning space $\mathcal{P}$, which is the output of the CLIP text encoder.
Instead of altering any weights in SD models, they optimize $v^*$ alone using \eref{eq:sd}.
To create images of personalized concepts, the inclusion of $S^*$ in the prompts (\textit{e.g.} a photo of $S^*$ dog) is the only required step.

\subsection{Multi-Stage Textual Inversion}
In some cases, TI fails to sufficiently represent the concept due to the inherent capacity limitations associated with using a single embedding token.
Moreover, this single embedding strategy is inappropriate for accommodating the changing process of diffusion models.
As explored in \citet{balaji2022ediffi, choi2022perception}, diffusion models display intriguing temporal dynamics throughout the process, necessitating different capacities at various diffusion steps.
In light of this, managing all denoising timesteps with a single embedding potentially has limitations due to the spectrum of local to global expressions embodied in paintings.
Thus, articulating paintings is intricately related to the denoising timesteps, which operate in a coarse-to-fine synthesis manner~\cite{balaji2022ediffi}. 
To address these challenges, we introduce a \textit{multi-stage TI} that employs multiple embeddings, each corresponding to specific diffusion stages (\fref{fig:model_overview}).

We first propose an extended textual embedding space $\mathcal{S}$.
The premise of the $\mathcal{S}$ space is to decompose the entire diffusion process into multiple distinct \textit{stages}.
To implement this, we split the denoising timesteps into $T$ chunks and denote each chunk as a stage.
Based on the $\mathcal{S}$ space, the multi-stage TI prepares the copies of the initial style token ($S^*$) as a multi-stage token set $\mathbf{S^*} = \{S^*_1, \dots, S^*_T\}$. In this way, the multi-stage TI projects a style image into $T$ style tokens, contrasting the TI that embeds it into a single token.
The token set is then encoded by a CLIP text encoder to form stage-wise embedding vectors, denoted as $\mathbf{v^*} = \{v^*_1, \dots, v^*_T\}$.
Lastly, the multi-stage TI optimizes these embeddings following the subsequent equation.
\begin{equation}
\mathbf{v^*} = \argmin_{\mathbf{v}}\mathbb{E}_{z, \mathbf{v}, \epsilon, t} [||\epsilon - \epsilon_{\theta}(z_t, t, c(v_t))||^2_2].
\label{eq:sd_msti}
\end{equation}

The application of multi-stage TI significantly enhances the representation capacity beyond that of vanilla TI, which we will illustrate in a series of experiments.
Furthermore, this method enables the fusion of multiple tokens, each originating from different styles, at a specific stage $t$.
Consequently, it facilitates the creation of unique and novel styles tailored to the user's individual preferences.

\subsection{Context-Aware Text Prompt}
While the multi-stage TI enhances representational capacity, it still faces fundamental problems when training with a style reference; the style and context of the image may become entangled during the optimization of the embeddings.
This problem mainly arises from attempts to encapsulate all features of the image into $S^*$, not just the style aspect.
As depicted in \fref{fig:prompt_preanalysis}, without contextual information in the training prompt, the model overlooks the context of inference prompt.
However, when we inject contextual descriptions into the training prompt, the model better disentangles the style from the context.
In our observations, such a phenomenon occurs more frequently as the representational capacity increases, likely due to the model's increased efforts to accommodate all information within its capacity.

Hence, we construct training prompts to include contextual information about the style image.
Let $C=[C_o, \mathbf{S^*}]$ be the vanilla prompt used in multi-stage TI training, where $C_o$ is the opening text (\textit{e.g.} ``a painting"), and $\mathbf{S^*}$ is multi-stage style token set, described above.
In the proposed strategy, we incorporate a contextual descriptor $C_c$ (\textit{e.g.} ``of a woman in a blue dress") into the middle of the prompt (\fref{fig:model_overview}), \textit{i.e.} $C=[C_o, C_c, \mathbf{S^*}]$.
We annotate all the non-style attributes (\textit{e.g.} objects, composition, and background) from the style image to form the contextual descriptor.
When we caption non-style attributes, BLIP-2~\cite{li2023blip} is employed to aid in the automatic prompt generation.

Although a context-aware prompt significantly reinforces style-context decoupling, for some style images with complicated contexts (\fref{fig:prompt_preanalysis}), BLIP-2 might not capture all details, which could limit the model's disentanglement capability.
In such cases, we further refine caption $C_c$ based on human feedback (e.g., caption by humans).
This human-in-the-loop strategy is straightforward yet markedly improves the model's ability to disentangle styles. 
Since our goal is one-shot model training, the time spent refining the caption is minimal; typically less than a minute. 
With the context-aware prompt, the text-to-image models can now distinguish style elements from contextual ones and specifically embed these into the (multi-stage) style embeddings $\mathbf{v}^*$.
The motivation for augmenting the training prompt is also suggested in StyleDrop~\cite{sohn2023styledrop}, a current personalization approach in the text-to-image diffusion model.

\figTIComp{}
\figSTComp{}

\subsection{Style and Context Guidance}
Classifier-free guidance~\cite{ho2022classifier} improves conditional image synthesis.
It samples adjusted noise prediction $\hat{\epsilon}(.)$, by leveraging unconditional output under null token $\emptyset$ as: $\hat{\epsilon}(\mathbf{v}) = \epsilon(\emptyset) + \lambda(\epsilon(\mathbf{v}) - \epsilon(\emptyset))$, where, $\lambda$ is the guidance scale and we omit $c(.)$, $z$ and $t$ for brevity.

In style-guided image synthesis, this guidance pushes both style and context uniformly with $\lambda$.
The uniform guidance could face limitations since the spectrum of ``style" of artistic paintings is wider than that of natural photos.
Given this diversity, a more nuanced control mechanism is required.
Furthermore, there exist demands to individually control style and context in the art-making process.
To this end, we propose style and context guidance as in below.
\begin{align}
\hat{\epsilon}(\mathbf{v}) = \epsilon(\emptyset) 
&+ \lambda_s[\epsilon(\mathbf{v}) - \epsilon(\mathbf{v_c})]
+ \lambda_c[\epsilon(\mathbf{v_c}) - \epsilon(\emptyset)] \nonumber\\
&+ \lambda_c[\epsilon(\mathbf{v}) - \epsilon(\mathbf{v_s})] 
+ \lambda_s[\epsilon(\mathbf{v_s}) - \epsilon(\emptyset)]
\label{eq:guidance}
\end{align}
where, $\mathbf{v}_s, \mathbf{v}_c$ are the embeddings of prompts $C, C_c$, respectively.
$\lambda_s, \lambda_c$ denote style and context guidance scale.
We derive \eref{eq:guidance} by decomposing $\mathbf{v}$ into $\mathbf{v}_s, \mathbf{v}_c$.
We employ two paired terms to balance the influence of each guidance.
Please refer to Suppl. for detailed derivation and analysis.

By separating the guidance into style and context, users are afforded the flexibility to control these elements individually.
Specifically, an increase in $\lambda_c$ increases the model's sensitivity towards context (\textit{e.g.} inference prompt or content image), whereas amplifying $\lambda_s$ leads the model towards a more faithful style reproduction.
This flexible design allows users to generate stylistic output tailored to their individual preferences, and it also facilitates the adoption of various styles, each with a range of complexities~\cite{hong2023aespanet}.

\subsection{Style Transfer}
\ours\ transmits styles by inverting a content image into a noisy sample and then denoising it towards the style domain~\cite{meng2021sdedit}.
With this approach, however, the preservation of content would be suboptimal~\cite{ahn2023interactive}.
To improve this, we inject additional conditions from the content image into the model~\cite{zhang2023adding} (\fref{fig:model_overview}).
This straightforward pipeline well preserves with the structure of the content image, while effectively replicating styles.
Moreover, by leveraging a powerful prior knowledge from text-to-image models, the style quality of \ours\ surpasses that of traditional methods.
\figTICompTradeoff{}
\tableTIComp{}
\figSOComp{}
\tableSTComp{}

\section{Experiment}
\noindent\textbf{Implementation details.} We use $T=6$ for multi-stage TI and utilize human feedback-based context prompts by default. Please refer to Suppl. for more details.

\smallskip
\noindent\textbf{Datasets.} We collected a set of 32 images representing various artistic styles, following the literature on style transfer~\cite{artgan2018}.
To evaluate text-to-image synthesis, we prepared 40 text prompts, as described in Suppl.

\smallskip
\noindent\textbf{Baselines.} In terms of text-to-image synthesis, we compare \ours\ against diffusion-based personalized methods, ranging from textual inversion to model-optimization approaches.
For the style transfer task, we compare our method to state-of-the-art style transfer frameworks.
We utilize official codes for all the methods used in the comparison.

\smallskip
\noindent\textbf{Evaluation.}
Text and image scores, based on CLIP, measure the alignment with a given text prompt and style image, respectively.
Style score assesses the style consistency by calculating the similarity of Gram features between the style and generated images.
More details are provided in Suppl.

\figMSTI{}
\figMSAnalysis{}

\subsection{Style-Guided Text-to-Image Synthesis}
\tref{table:ti_comp} and \fref{fig:ti_comp_tradeoff} show quantitative results.
\ours\ delivers a robust performance while managing the trade-off between text and style scores.
A tendency is noted that an overemphasis on input text prompts may lead to a compromise in style quality.
Despite this, \ours\ effectively balances these aspects, yielding a performance that goes beyond the trade-off line, indicative of outstanding capability.
User score also supports the distinction of \ours.

As shown in \fref{fig:ti_comp}, previous inversion-based methods (TI, InST, and XTI) effectively preserve the context of text prompts but fall short in adopting the intrinsic artwork of style images.
Conversely, the model optimization-based methods (DreamBooth, CustomDiffusion) excel in delivering styles but struggle to adhere to the prompt or introduce objects in style images (3rd row).
\ours, in contrast, not only faithfully follows text prompts but also accurately reflects the delicate artistic features of style images.

\tableAblation{}

\subsection{Style Transfer}
As an extended application, \ours\ also conducts style transfer.
As shown in \tref{table:st_comp}, we quantitatively compare with previous style transfer studies.
Note that since most prior studies have employed Gram loss to boost style quality, we report a CLIP-based image score as an evaluation metric for a more fair comparison.
In this benchmark, \ours\ achieves state-of-the-art performance across text and image scores as well as user preference.
\fref{fig:st_comp} also provides evidence of \ours's effectiveness.
Our method adeptly captures style features such as polygon shapes or subtle brushwork present in style images.
These results highlight the method's capacity to accurately mirror both the thematic intent and the stylistic nuances of the source artwork.

\figPromptAnalysis{}

\subsection{Stylize My Own Object in My Own Style}
Beyond style transfer that stylizes \textit{my image}, one might desire to stylize \textit{my object}~\cite{sohn2023styledrop}.
In such a scenario, a user leverages both their object and style images.
As \ours\ employs an inversion-based approach, this can be readily accomplished by simply training an additional embedding for the object.
Subsequently, the user freely merges style and object tokens in the inference prompt to generate images.
As depicted in \fref{fig:so_comp}, \ours\ excels in accurately reflecting both the style and object

\figGuidanceAnalysis{}
\figStyleMixing{}

\subsection{Model Analysis}
\noindent\textbf{Ablation study.}
In \tref{table:ablation}, we evaluate each component of our method.
The usage of multi-stage TI substantially augments both the text and style score, with a marked increase in style quality, accentuating the pivotal role of this module in creating artistic stylization products.
A context-aware prompt yields a modest alteration in the quantitative metrics, yet provides a considerable contribution to the qualitative, which we will discuss in the following section.
Style and context (S\&C) guidance considerably impacts scores, reinforcing its significance in sustaining the comprehensive quality and coherence of the generated outputs.

\smallskip
\noindent\textbf{Multi-stage TI.} In \fref{fig:ms_ti}, we delve into the influence of the number of stages ($T$) on performance.
A transition from $T=1$ to $4$ results in substantial improvement.
Upon reaching $T=6$, the performance begins to navigate trade-off contours, prompting us to select $T=6$ for the final model, as we seek to improve the text alignment of the synthesized images.
Nevertheless, users have the flexibility to choose a different $T$ value according to their preference.
In \fref{fig:ms_analysis}, we provide a visual comparison of the outcomes when $T$ is set to 1, 2, and 6.
While $T=1$ struggles to reflect the artistic features of the style image or comprehend the input prompt, $T=2$ uplifts the quality, yet it also falls short of embracing the style.
In contrast, $T=6$ proves proficient at mimicking the style image, effectively replicating delicate brushwork (1st row) or emulating the pointillism style (2nd row).

\smallskip
\noindent\textbf{Context-aware prompt.} \fref{fig:prompt_analysis} presents a visual comparison of three prompt constructions.
Training the model without any contextual description (\textit{i.e.} using ``A painting in $S^*$ style.") poses a significant challenge, as it struggles to distinguish style from the context within the style image.
Subsequently, this often results in the generation of elements that exist in the style reference, such as objects or scene perspective.
The introduction of a contextual prompt considerably alleviates this issue, aiding the model in better separating stylistic elements from context.
However, the automatic prompt construction does not fully resolve this, as BLIP-based captions often fail to capture all the details of the style image.
The most effective solution is leveraging human feedback in the construction of prompts.
This approach effectively tackles the issue, resulting in a more robust separation of style and context in the generated outputs.

\smallskip
\noindent\textbf{Guidance.} In \fref{fig:guidance_analysis}, we explore style and context guidance by adjusting the scale parameters.
When we amplified the style guidance strength ($\lambda_s$), the model mirrors the style image, illustrating style guidance's capability in managing the image's aesthetics.
Yet, overemphasis on style risks compromising the context, leading to outputs that, while stylistically congruent, might diverge from the intended context.
On the other hand, strengthening context guidance ($\lambda_c$) ensures the output resembles the inference prompt, highlighting context guidance's essential role in preserving contextual integrity.
However, excessively strong context guidance could steer the output away from the original style, underlining the need for a nuanced balance of guidance for generating visually appealing and contextually accurate images.
Nevertheless, this offers a new dimension of control over the synthesized image, differing from the classifier-free guidance~\cite{ho2022classifier}. 
The additional control is a crucial element in the workflow of digital art production, considering its delicate and nuanced final outcomes.

\smallskip
\noindent\textbf{Style mixing.} As shown in \fref{fig:style_mixing}, multi-stage TI opens up a novel avenue for an intriguing aspect of style mixing from diverse style references.
This process empowers users to customize a unique style by deploying different style tokens at each stage $t$.
The style tokens close to $t=T$ predominantly impact the structure of the image, akin to broad strokes, while tokens closer to $t=0$ affect local and detailed attributes, akin to intricate brushwork.
To provide a concrete point of comparison, we present a baseline model that incorporates all style tokens at every stage, using the prompt ``A painting in $S^A_t$, $S^B_t$, $S^C_t$ styles".
While the baseline produces reasonable style quality, it lacks a control factor for extracting partial stylistic features from the reference.
Consequently, the fusion of styles with multi-stage TI underscores the creative and flexible nature of our model, offering users a broad range of applications for artistic creation.

\section{Conclusion}
We have introduced \ours, a novel image generation method with a given style reference.
By optimizing multi-stage TI with a context-aware text prompt, \ours\ achieves remarkable performance in both text-to-image synthesis and style transfer.
Content and style guidance provides a more adaptable way of handling diverse style references.

\smallskip
\noindent\textbf{Limitations.} While \ours\ exhibits outstanding ability in generating artistic imagery, it is important to acknowledge its limitations within the intricate context of artistic expression.
The vast spectrum of artistry, spanning from primitive elements to more nuanced and abstract styles (such as surrealism), demands thorough definition and examination from both artistic and technological perspectives.
\bibliography{}

\appendix
\newpage
\twocolumn[
\begin{@twocolumnfalse}
\begin{center}
\textbf{\LARGE Appendix}
\vspace{2em}
\end{center}
\end{@twocolumnfalse}
]
\section{Style and Context Guidance}
In this section, we derive how we obtain the style and context guidance depicted in \eref{eq:guidance} in the main text.

\smallskip
\noindent\textbf{Derivation.} Classifier-free guidance~\cite{ho2022classifier} modifies the predicted noise estimation to include the gradient of the log-likelihood of $p(\mathbf{v} | x_t)$, where $\mathbf{v}$ denotes the conditional embedding tokens (of a given text) and $x_t$ is the denoised sample at $t$-th denoising step.
Given that $p(\mathbf{v} | x_t) \propto p(x_t | \mathbf{v}) / p(x_t)$, it follows that $\nabla_{x_t} \log p(x_t | \mathbf{v}) \propto \nabla_{x_t} \log p(x_t, \mathbf{v}) - \nabla_{x_t} \log p(x_t)$.
We parameterize the exact score with the score estimator as $\epsilon_\theta(.) \propto \nabla_{x_t} \log p(.)$, enabling us to derive the classifier-free guidance term as:
\begin{equation}
\hat{\epsilon}_\theta(x_t, \mathbf{v}) = \epsilon_\theta(\emptyset) 
+ \lambda_n[\epsilon_\theta(x_t, \mathbf{v}) - \epsilon_\theta(x_t, \emptyset)].
\end{equation}

To derive the proposed style and context guidance, we first decompose text conditional embedding tokens $\mathbf{v}$ into its style and context components as $\mathbf{v} = \mathbf{v_s} \cap \mathbf{v_c}$.
As an example, $\mathbf{v}$ is the embedding tokens of inference prompt ``A painting of a house in the style of $S^*$", while $\mathbf{v_s}$ and $\mathbf{v_c}$ are the embedding tokens of ``A painting of a house" and ``in the style of $S^*$" prompts, respectively.
Given these partitions, we can rewrite $p(\mathbf{v}|x_t)$ using the terms below:
\begin{align}
p(\mathbf{v}|x_t) &= p(\mathbf{v_s} \cap \mathbf{v_c} | x_t) \\
                  &\propto p(\mathbf{v_c} | x_t)p(\mathbf{v_s}|\mathbf{v_c}, x_t)\label{eq:suppl_p_first}\\
                  &\propto p(\mathbf{v_s} | x_t)p(\mathbf{v_c}|\mathbf{v_s}, x_t).\label{eq:suppl_p_second}
\end{align}
From \eref{eq:suppl_p_first}, we derive the following expression:
\begin{equation}
p(\mathbf{v}|x_t) \propto \frac{p(x_t | \mathbf{v_c})}{p(x_t)}\frac{p(x_t | \mathbf{v_s}, \mathbf{v_c)}}{p(x_t|\mathbf{v_c})}
\end{equation}
As in \citet{ho2022classifier}, we deduce the above equation to the style and context guidance as in below.
\begin{equation}
\begin{aligned}
\hat{\epsilon}^1_\theta(x_t, \mathbf{v}) &= \epsilon_\theta(x_t, \emptyset) \\
&+ \lambda_c[\epsilon_\theta(x_t, \mathbf{v_c}) - \epsilon_\theta(x_t, \emptyset)] \\
&+ \lambda_s[\epsilon_\theta(x_t, \mathbf{v}) - \epsilon_\theta(x_t, \mathbf{v_c})]
\label{eq:suppl_guidance_first}
\end{aligned}
\end{equation}
Similarly, using \eref{eq:suppl_p_second}, we derive another guidance term:
\begin{equation}
\begin{aligned}
\hat{\epsilon}^2_\theta(x_t, \mathbf{v}) &= \epsilon_\theta(x_t, \emptyset) \\
&+ \lambda_s[\epsilon_\theta(x_t, \mathbf{v_s}) - \epsilon_\theta(x_t, \emptyset)] \\
&+ \lambda_c[\epsilon_\theta(x_t, \mathbf{v}) - \epsilon_\theta(x_t, \mathbf{v_s})]
\label{eq:suppl_guidance_second}
\end{aligned}
\end{equation}

\tableSupplSetting{}

Rather than relying solely on either \eref{eq:suppl_guidance_first} or \eref{eq:suppl_guidance_second}, we employ a balanced guidance approach by integrating both forms of guidance as shown in \eref{eq:guidance} in the main text.
We will elaborate on the benefits of utilizing both terms in the following section.
In practice, alongside \eref{eq:guidance}, we also leverage the conventional classifier-free guidance.
Therefore, the final style and context guidance is as follows:
\begin{equation}
\begin{aligned}
\hat{\epsilon}_\theta(x_t, \mathbf{v}) &= \epsilon_\theta(x_t, \emptyset) \\
&+ \lambda_n[\epsilon_\theta(x_t, \mathbf{v}) - \epsilon_\theta(x_t, \emptyset)] \\
&+ \lambda_c[\epsilon_\theta(x_t, \mathbf{v_c}) - \epsilon_\theta(x_t, \emptyset)] \\
&+ \lambda_s[\epsilon_\theta(x_t, \mathbf{v}) - \epsilon_\theta(x_t, \mathbf{v_c})] \\
&+ \lambda_s[\epsilon_\theta(x_t, \mathbf{v_s}) - \epsilon_\theta(x_t, \emptyset)] \\
&+ \lambda_c[\epsilon_\theta(x_t, \mathbf{v}) - \epsilon_\theta(x_t, \mathbf{v_s})]
\end{aligned}
\end{equation}

By utilizing both the proposed and the classifier-free guidances, we can subtly adjust the scaling of style and context guidance ($\lambda_s, \lambda_c$), starting from a zero value.
For instance, to minimize the influence of the style guidance on the output images, one merely needs to set $\lambda_s$ to zero without changing other hyperparameters, and similarly, this applies to the context guidance as well.
Without the incorporation of classifier-free guidance, users would be exhausted from initiating the guidance scaling search from scratch.
We describe the guidance scaling parameters in \tref{table:suppl_settings}.

\section{Experimental Settings}
\noindent\textbf{Implementation details.} \tref{table:suppl_settings} demonstrates hyperparameters used for our method. We optimize \ours\ with \eref{eq:sd_msti} (in the main text), which is a similar training procedure to textual inversion~\cite{gal2022image}.
We utilize a single A100 GPU for both optimization and inference.

For the style transfer task, we extract a depth map condition from a content image and then encode structure features by adopting ControlNet~\cite{zhang2023adding}.
Although various conditional modalities can be considered, we observed that a depth map is the most suitable choice for style transfer.
An edge map tends to strongly preserve the structure of content images so that the structural styles are not effectively conveyed to output images.
While one might consider employing a segmentation map to better preserve the structure, this approach can introduce artifacts, especially if the segmentation map is not flawlessly generated.

\smallskip
\noindent\textbf{Datasets.}
To build a dataset used in our model evaluation, we collect most of the artistic paintings from the WikiArt dataset~\cite{artgan2018}, while some of the modern art and illustration are from Unsplash\footnote{https://unsplash.com}, licensed free images only.
In \tref{table:suppl_ex_prompts}, we show inference prompts used in the evaluation.

\smallskip
\noindent\textbf{Evaluation.}
For model evaluation, we employ CLIP-based text and image scores as well as Gram-based style score. 
The CLIP text score evaluates the alignment of a generated image $I$ with a given text prompt $C$ as in below.
\begin{equation}
\text{TextScore}(I, C) = max(100 * cos(E_I, E_C), 0), \nonumber
\end{equation}
where, $E_I$ denotes visual CLIP embedding of image $I$ and $E_C$ is textual CLIP embedding of prompt $C$.
The CLIP image score measures the alignment of a generated image $I$ and a given style reference image $S$ with the following equation.
\begin{equation}
\text{ImageScore}(I, S) = max(100 * cos(E_I, E_S), 0). \nonumber
\end{equation}

The Gram-based style score is a widely used metric in style transfer.
It quantifies the textural alignment between two images using Gram features, as below equation.
\begin{equation}
\text{StyleScore}(I, S) = 50 - \frac{1}{L}\sum^L_{l=1}\frac{1}{\mathcal{B}}\sum_{\forall (i,j)\in\mathcal{B}} cos(P^i_I, P^j_S) \nonumber
\end{equation}
Here, $L$ is the number of layers of VGG16~\cite{simonyan2014very}, which is used to calculate Gram features. $\mathcal{B}$ represents the selected patch pair set and $P^i$ is $i$-th image patch of an image.
Notably, we assess the style alignment patch-wise; this technique is more adept at capturing local style features than when analyzing the entire image.
Additionally, to align the magnitude of this score with others, we subtract Gram similarity to 50.
For an efficient implementation, we choose five patches, each of size 224$\times$224, by cropping from the four corners and the center of the image.

\smallskip
\noindent\textbf{User study.}
We asked 37 participants where they were asked to select the best results based on how closely the outputs adhered to the style of reference images and the context of inference prompts (or content images).
Each participant was tasked to vote on 9 questions for text-to-image synthesis (yielding 333 responses in total) and 7 questions for style transfer (yielding 259 responses in total).
They were presented with style images, inference prompts, and the results from \ours\ to other methods.
For the style transfer evaluations, content images were displayed in place of the inference prompts.
The questionnaires used in the user study are detailed in \tref{table:suppl_user_study}.
We determined the User score based on the ratio of instances voted as the best.

\figSupplFewshot{}

\section{Additional Model Analysis}
\noindent\textbf{Training/inference time.}
Textual inversion methods typically demand increased training time due to the forward/backward passes in the CLIP text encoder.
Nevertheless, the inference time difference with the model optimization-based approach is minimal.
In our measurement, with a batch size of one, DreamBooth, TI, and \ours\ requires 300s, 620s, and 580s, respectively.
With 8 batch size, DreamBooth, TI, and \ours\ takes 60s, 500s, and 480s.
Despite the additional time required, \ours\ proves its worth by delivering superior stylization outcomes as shown in a series of experiments.

\smallskip
\noindent\textbf{Few-shot text-to-image synthesis.}
\fref{fig:suppl_fewshot} depicts the change in performance trade-offs for diffusion-based personalization methods when transitioning from a one-shot to a few-shot training regime.
For this analysis, all methods are trained using five artistic-style images.
Importantly, model optimization-based frameworks, such as DreamBooth and CustomDiffusion, exhibit marked improvements in text scores (as compared to \fref{fig:ti_comp_tradeoff}, in the main text) due to their capability derived from directly optimizing parameters of denoising UNet.
Conversely, textual inversion-based methods, such as TI and XTI, fail to leverage the benefits of multiple images due to the inherent capacity constraint of the embedding optimization nature.
Despite \ours\ employing a textual inversion-based approach, our method not only distinguishes from other textual inversion frameworks but also achieves a trade-off performance comparable to CustomDiffusion, attributed to the timestep-aware textual embedding and context-aware training prompt strategy.

\smallskip
\noindent\textbf{Style and context guidance.}
In \fref{fig:suppl_abl_guidance}, we compare three different guidance settings: 1) $\hat{\epsilon}_c(\mathbf{v})$, shown in \eref{eq:guidance} in the main text, 2) $\hat{\epsilon}^1_\theta(\mathbf{v})$, shown in \eref{eq:suppl_guidance_first}, and 3) $\hat{\epsilon}^2_\theta(\mathbf{v})$, shown in \eref{eq:suppl_guidance_second}.
Note that the difference in the outputs ($\epsilon(\mathbf{v})$) relative to the one with null condition ($\epsilon(\emptyset)$) exceeds those of with style-only ($\epsilon(\mathbf{v_s})$) or context-only ($\epsilon(\mathbf{v_c})$).
This is because the null conditioned outputs are presumed to produce arbitrary imagery.
Hence, when we only rely on $\hat{\epsilon}_\theta^1(\mathbf{v})$ or $\hat{\epsilon}_\theta^2(\mathbf{v})$, the guidance influence between style and context might be imbalanced even with the same scaling parameters, $\lambda_s, \lambda_c$. For instance, with the $\hat{\epsilon}_\theta^1(\mathbf{v})$ guidance, the computation involves differences between $\epsilon(\mathbf{v_c})$ and $\epsilon(\emptyset)$, which amplifies contextual details at the expense of style attributes. On the other hand, $\hat{\epsilon}_\theta^2(\mathbf{v})$ emphasizes style, often overlooks the contextual facets. To achieve a harmonious blend of style and context guidance, we incorporate both guidance forms.

\figSupplAblGuidance{}

\begin{figure}[t]
\centering
\includegraphics[width=1.0\linewidth]{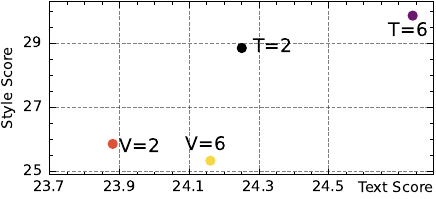}
\caption{\textbf{Multi-stage TI ($T$) vs. naive multi-token ($V$).}}
\label{fig:t_vs_v}
\end{figure}

\smallskip
\noindent\textbf{Ablation study on multi-stage TI.}
To demonstrate the necessity of multi-stage TI, we compare our proposed TI with a naive multi-token approach in \fref{fig:t_vs_v}.
Naive multi-token TI ($V=2,6$) is the method in which only the number of embedding tokens is increased.
Our multi-stage TI approach outperforms both StyleScore and TextScore.

\smallskip
\noindent\textbf{The effectiveness of \# T.}
We observed a saturation of StyleScore when $T \geq 6$ (\fref{fig:ms_ti}).
This indicates that more embeddings beyond this threshold do not contribute new stylistic nuances and may instead introduce redundancy in artistic expression.
For TextScore, we observed an unexpected trend at $T \geq 6$; it starts to generate subjects not specified in the prompt, but presented in style image.
We hypothesize that as $T$ increases to a high value, the embeddings try to learn context of style image, beyond stylistic focus; a potential overextension of the embedding capacity.
Comparatively, DreamBooth's approach, which involves modifying a large set of parameters, manifests a substantially lower TextScore.
Thus, we speculate that increasing $T$ to an extremely high value may yield trends similar to DreamBooth.
Nevertheless, regarding TextScore, we believe further investigation is required, and we are grateful for the highlight of this aspect, which had previously eluded our consideration.

\smallskip
\noindent\textbf{When the style guidance is effective?}
In \fref{fig:suppl_sg}, we investigate when the style guidance significantly influences the resulting synthesized image.
Our observations indicate that style guidance plays a crucial role in ensuring proper style expression, especially when the output, with zero style guidance ($\lambda_s=0$), fails to capture the stylistic nuances present in a reference image.
Conversely, when the output already effectively expresses the stylistic elements, the influence of style guidance becomes marginal, even with a substantial increase in the scaling parameters $\lambda_s$.

Given these observations, it is now essential to distinguish when the model can accurately replicate style images without the need for style guidance.
We hypothesize that style pattern repeatability~\cite{hong2023aespanet} is linked to this ability to some extent.
Specifically, when a stylistic feature is complex, the model struggles to capture all its stylistic details.
In such cases, the style guidance serves as an effective booster for the model's style representation.
Another influencing factor, we conjecture, is the global colortone of the style image.
We observed that when there is a pronounced disparity in the colortone of the style image compared to a natural photo, the model often struggles to mimic it, even if the style is not particularly complicated.
Despite our observations, determining whether a model can easily adapt to a particular style remains a non-trivial question, and we leave this investigation for future research.

\smallskip
\noindent\textbf{When the context guidance is effective?}
\fref{fig:suppl_cg} demonstrates scenarios in which the context guidance proves beneficial for artistic image synthesis.
We hypothesize that the context guidance is particularly effective when the style image is expressed in a very abstract manner, especially when the structures of the desired subjects to draw deviate significantly from the stylistic expression of the reference image.
Conversely, when the style images aptly capture the details of some subjects (as realism paintings do usually), the model can render the desired subject in that style without needing context guidance.
However, determining when the personalized diffusion model can adeptly convey the subject with a given style image remains an open question, and we leave this exploration for future work.

\smallskip
\noindent\textbf{Style transfer}
In \fref{fig:suppl_abl_style_transfer}, we illustrate the role of condition modality in the style transfer task.
The figure indicates that when the additional condition modality is incorporated via ControlNet~\cite{zhang2023adding}, the resulting style-transferred images retain a significant degree of fidelity to the original contents' structure.
On the other hand, methods that bypass this step and rely solely on the image inversion technique~\cite{meng2021sdedit} often introduce considerable alterations to the initial structure.

\figSupplAblStyleTransfer{}

\smallskip
\noindent\textbf{Training progress.}
In \fref{fig:suppl_training_A},\ref{fig:suppl_training_B}, and \ref{fig:suppl_training_C}, we compare model optimization- and textual inversion-based methods throughout their training progress.
To conduct this, we double the training steps and plot the intermediate results of each method.
This allows us to examine the training tendencies and to determine whether the methods fall into overfitting.
Through this inspection, we highlight the strengths and weaknesses of these approaches.
The model optimization-based approach, DreamBooth~\cite{ruiz2023dreambooth}, CustomDiffusion~\cite{kumari2023multi}, exhibits superior style adaptation capability owing to its rich capacity as we train the model directly.
However, this approach is prone to overfitting.
When overfitting occurs, the model tends to generate images that are very similar to the style images, disregarding all the context from the inference prompt.
A critical point to note is that the training steps at which overfitting occurs vary significantly to style images.
In practice, this is problematic because users cannot select a fixed number of training steps.
Instead, they need to inspect all intermediate samples to identify the best checkpoint, which is highly time-consuming and unsustainable in real-world production.
Conversely, the textual inversion-based approach~\cite{gal2022image} might not replicate the style image as effectively as model optimization does, but it is less prone to overfitting.
This tendency might appear to be symptomatic of underfitting.
\ours\ takes the strengths of both approaches while mitigating their weaknesses.
It avoids overfitting in the same way as the textual inversion-based method, yet it also adeptly captures the stylistic nuances found in the style image, akin to the model optimization-based approach.
This emphasizes the superiority and practicality of \ours.

\begin{figure}[t]
\centering
\includegraphics[width=1.0\linewidth]{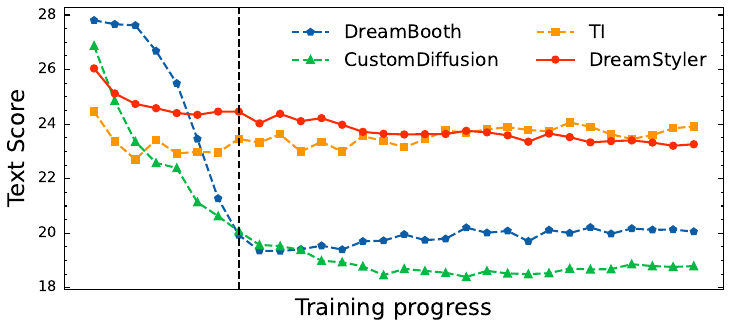}
\caption{\textbf{Training progress.}}
\label{fig:overfitting}
\end{figure}

In \fref{fig:overfitting}, we demonstrate quantitative results of such overfitting phenomena.
In our task, overfitting is characterized by the model's neglect of the text prompt and verbatim reproduction of the style image and this can be identifiable through changes in the TextScore.
Notably, CustomDiffusion and DreamBooth exhibit a sudden and substantial drop in TextScore at the point when the results are merely copied from style images.
On the contrary, \ours\ performs better in mitigating overfitting than DreamBooth.

\section{Additional Qualitative Results}
In this section, we provide additional qualitative results for better comparisons with previous studies.
As shown in \fref{fig:suppl_ti_comp}, \ours\ successfully generates results with given prompts while depicting the specific style of reference images such as pointillism or watercolor.
We also provide additional comparison results on the style transfer task in \fref{fig:suppl_st_comp}.
\ours\ transfer the style impeccably while preserving the semantic information of content images.

\figSupplSG{}
\figSupplCG{}

\figSupplTrainingA{}
\figSupplTrainingB{}
\figSupplTrainingC{}

\figSupplTIComp{}
\figSupplSTComp{}
\tableExPrompts{}
\tableUserStudy{}

\end{document}